\documentclass{article}

\usepackage{float}
\usepackage[table]{xcolor}
\newcommand{\bestcolor}{\cellcolor[HTML]{c2d2ee}}
\newcommand{\secondcolor}{\cellcolor[HTML]{ebf2ff}}
\newcommand{\bestcolortext}[1]{\colorbox[HTML]{c2d2ee}{#1}}
\newcommand{\secondcolortext}[1]{\colorbox[HTML]{ebf2ff}{#1}}

\newcommand{\blueshaded}[1]{\colorbox[HTML]{dfe5f2}{#1}}
\newcommand{\orangeshaded}[1]{\colorbox[HTML]{f9ede5}{#1}}
\newcommand{\greenshaded}[1]{\colorbox[HTML]{edf4e6}{#1}}

\usepackage{color}
\newcommand{\blue}[1]{\textcolor{blue}{#1}}

\usepackage{lipsum} 
\usepackage{wrapfig} 
\usepackage{booktabs} 
\newcommand\independent{\protect\mathpalette{\protect\independenT}{\perp}}
\def\independenT#1#2{\mathrel{\rlap{$#1#2$}\mkern2mu{#1#2}}}

\usepackage[preprint]{neurips_2025}

\usepackage[utf8]{inputenc} 
\usepackage[T1]{fontenc}    
\usepackage{hyperref}       
\usepackage{url}            
\usepackage{booktabs}       
\usepackage{amsfonts}       
\usepackage{nicefrac}       
\usepackage{microtype}      
\usepackage{xcolor}         

\usepackage{microtype}
\usepackage{graphicx}
\usepackage{subfigure}
\usepackage{booktabs} 

\usepackage{amssymb}
\usepackage{pifont}

\usepackage{enumitem}
\usepackage{amsmath}
\usepackage{amssymb}
\usepackage{mathtools}
\usepackage{amsthm}

\usepackage[capitalize,noabbrev]{cleveref}
\theoremstyle{plain}

\theoremstyle{definition}

\theoremstyle{remark}

\usepackage[textsize=tiny]{todonotes}

\newlength\savedwidth

\newlength\savewidth
\newcommand\shline{\noalign{\global\savewidth\arrayrulewidth
                            \global\arrayrulewidth 1.5pt}%
                   \hline
                   \noalign{\global\arrayrulewidth\savewidth}}

\newcommand{\real}{\mathbb{R}}

\newcommand{\mx}{\boldsymbol{X}}

\title{CaPulse: Detecting Anomalies by Tuning in to \\ the Causal Rhythms of Time Series}

\author{%
 Yutong Xia$^{1,2}$, 
 Yingying Zhang$^{2}$,
 Yuxuan Liang$^{3}$\thanks{Yuxuan Liang is the corresponding author of this paper. Email: yuxliang@outlook.com}, \\
 \textbf{Lunting Fan$^{2}$, 
 Qingsong Wen$^{4}$, 
 Roger Zimmermann$^{1}$} 
 \vspace{0.3em}\\
  \textsuperscript{\rm 1}National University of Singapore  \textsuperscript{\rm 2}Alibaba Group\\
  \textsuperscript{\rm 3}The Hong Kong University of Science and Technology (Guangzhou)
  \textsuperscript{\rm 4}Squirrel AI  
  \vspace{0.2em}\\
 \texttt{yutong.xia@u.nus.edu;
 congrong.zyy@alibaba-inc.com;
 yuxliang@outlook.com}\\
 \texttt{lunting.fan@taobao.com;
 qingsongedu@gmail.com;
 rogerz@comp.nus.edu.sg}\\
}

\begin{document}

\maketitle

\begin{abstract}
Time Series Anomaly Detection (TSAD) has garnered considerable attention across diverse domains, yet existing methods often fail to capture the underlying mechanisms behind anomaly generation. In addition, TSAD often faces several data-related inherent challenges, i.e., label scarcity, data imbalance, and complex multi-periodicity. In this paper, we leverage causal tools and introduce a new causality-based framework termed \textbf{CaPulse}, which ``tunes in'' to the underlying ``causal pulse'' of time series data to effectively detect anomalies. Concretely, we begin by building a structural causal model to decipher the generation processes behind anomalies. To tackle the challenges posed by the data, we propose Periodical Normalizing Flows with a novel mask mechanism and carefully designed periodical learners, creating a periodicity-aware, density-based anomaly detection approach. Extensive experiments on seven real-world datasets demonstrate that CaPulse outperforms existing methods, achieving AUROC improvements of 3\% to 17\%, with enhanced interpretability. 
\end{abstract}

\section{Introduction}\label{sec:intro}
Time Series Anomaly Detection (TSAD) has gained significant attention in recent years~\citep{darban2024deeplearningtimeseries} due to its applications across diverse domains such as network security~\citep{ahmed2016survey}, finance~\citep{takahashi2019modeling}, urban management~\citep{bawaneh2019anomaly}, and cloud computing services~\citep{ren2019time,chen2024lara}. Traditional TSAD methods, including one-class support vector machines~\citep{scholkopf2001estimating} and kernel density estimation~\citep{kim2012robust}, rely heavily on handcrafted features and struggle to handle high-dimensional time series data effectively. In contrast, Deep Learning (DL)-based approaches have recently emerged, significantly improving detection performance thanks to their powerful representation learning capabilities ~\citep{ruff2018deep,sabokrou2018adversarially,goyal2020drocc}.

Despite their promise, DL-based methods for TSAD face several key limitations. \textbf{\textit{Mechanistically}}, they often overlook the underlying patterns behind anomaly generation in time series data, leading to models that lack interpretability and exhibit limited generalizability. Causal inference~\citep{pearl2000models} provides a powerful platform for investigating the underlying causal systems, with successful integration in DL methods across various tasks~\citep{lv2022causality,zhao2024causality}. Specifically, by incorporating a \textit{causal perspective}, models can uncover the true factors driving anomalies, rather than relying solely on statistical dependencies or superficial correlations. Thus causal-based methods not only improve generalization and be more robust in Out-of-Distribution (OoD) scenarios~\citep{yang2022towards} but also significantly enhance interpretability, providing deeper insights into the root causes of anomalies. This is particularly essential for downstream tasks such as root cause analysis, where pinpointing the specific factor responsible for an anomaly is critical, such as identifying a server overheating or a hardware malfunction causing a cloud services system downtime~\citep{li2022causal}. 
Yet, there is still room for further exploration of causal-based methods for TSAD.

\begin{wrapfigure}{R}{0.5\textwidth}
\vspace{-2.7em}
  \includegraphics[width=\linewidth]{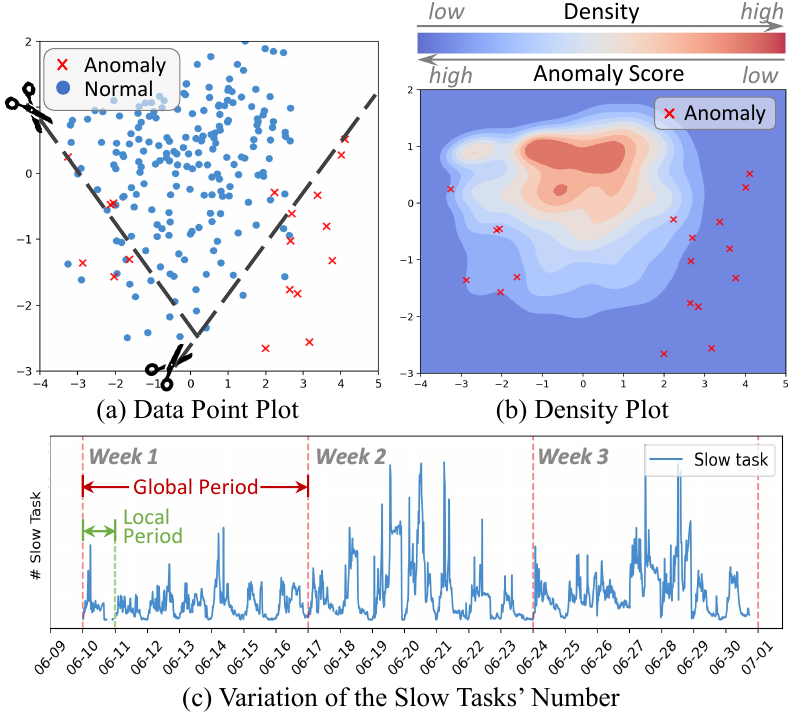}
  \vspace{-1.7em}
  \caption{(a) Data point plot and (b) density plot (c) multiple periodicities in the Cloud-S dataset.}\label{fig:intro}
  \vspace{-0.8em}
\end{wrapfigure}

In addition to the mechanical aspect, \textbf{\textit{intrinsically}}, TSAD is challenged by three characteristics in terms of data themselves: \textit{label scarcity}, \textit{data imbalance}, and \textit{multiple periodicities}. In practice, acquiring labeled anomalies is often impractical due to the significant manual effort and cost required~\citep{zhang2024advancing,chen2024cluster}. Even when labels are available, datasets typically consist of both normal and anomalous instances, resulting in overfitting to noisy labels~\citep{wang2019effective,huyan2021unsupervised} and degrading model performance~\citep{zhou2023detecting} (Figure~\ref{fig:intro}a). Additionally, time series exhibit multiple periodicities, with short-term cycles, e.g., hourly fluctuations, overlapping with long-term patterns that develop over weeks~\citep{wen2021robustperiod,wu2023timesnet}. We refer to them as \textit{local} and \textit{global} periodicities, respectively (Figure~\ref{fig:intro}c). This adds complexity to TSAD efforts: in cloud computing services, user misoperations often cause transient anomalies linked to short-term fluctuations, whereas long-term patterns typically signal machine failures.
However, existing TSAD methods fail to effectively address all three challenges simultaneously, underscoring the need for more advanced solutions.

In this paper, we first decipher the underlying generation process of anomalies and then provide solutions while addressing the challenges inherent to TSAD, i.e., label scarcity, data imbalance, and multiple periodicities. Specifically, we begin by adopting a causal standpoint, introducing a Structural Causal Model (SCM)~\citep{pearl2000models} to gain a deeper understanding of the causal mechanisms governing anomalies in time series. Building on this, we develop a novel DL framework that integrates causality-based solutions for accurate and interpretable TSAD. Meanwhile, motivated by the demonstrated success of density estimation in unsupervised anomaly detection~\citep{rudolph2021same,gudovskiy2022cflow,dai2022graphaugmented} (Figure~\ref{fig:intro}b), we develop a periodicity-aware, density-based approach that effectively addresses three inherent challenges in TSAD.
Our contributions are summarized as follows:
\begin{itemize}[leftmargin=*]
\vspace{-0.5em}
      \item \textbf{A causal view for TSAD.} 
      To uncover the underlying generation mechanisms driving anomalies, we present a causal view and propose an SCM for TSAD. Building on this, we leverage causal tools to introduce a new framework, \textbf{CaPulse}, which \textit{listens to the ``pulse'' of time series data -- its underlying ``causal'' rhythms -- and identifies when something is out of sync}. Like a capsule, CaPulse serves as an anomaly detector by pinpointing the true underlying issues in time series.
 \vspace{-0.2em}
      \item \textbf{A novel periodicity-aware density-based approach.} To tackle the intrinsic challenges of data, we propose Periodical Normalizing Flows to enable \textit{unsupervised density-based} anomaly detection. For capturing \textit{multi-period} dynamics, CaPulse integrates different periods' local information by learning causal pyramid representations as conditioning inputs, and global period information is incorporated via a novel mask mechanism.
 \vspace{-0.2em}
    \item \textbf{Comphrehensive empirical evidence.} We validate the effectiveness of CaPulse through extensive experiments on seven real-world datasets spanning five different domains. The results show that the proposed model consistently outperforms existing baselines on most datasets, achieving AUROC improvements ranging from 3\% to 17\%, while also providing clearer interpretability.
\end{itemize}

\vspace{-1em}
\section{Preliminaries}
\vspace{-0.5em}
\subsection{Problem Statement}
\vspace{-0.5em}
In this paper, we focus on \emph{unsupervised} anomaly detection in multivariate time series data. Let $\boldsymbol{X}^{1:T} = \{\boldsymbol{x}_1, \cdots, \boldsymbol{x}_T\} \in \mathbb{R}^{T \times D}$ represent multivariate time series, each $\boldsymbol{x}_t \in \mathbb{R}^D$ denotes the data at time point $t$, where $T$ is the length of the time series, and $D$ is the dimensionality. For a given $\boldsymbol{X}^{1:T}$, our target is to yield anomaly scores for all time points, denoted as $\boldsymbol{\tau}^{1:T} = \{\tau_1, \cdots, \tau_T\} \in \mathbb{R}^{T}$, where each $\tau_t \in \mathbb{R}$ indicates the likelihood of an anomaly at time $t$. For evaluation, we consider a corresponding set of labels $\boldsymbol{y}^{1:T} = \{y_1, \cdots, y_T\} \in \mathbb{R}^{T} $, where $y_t \in \left \{0, 1 \right \}$ indicates whether a time point is normal ($y_t = 0$) or anomalous ($y_t = 1$). For conciseness, we refer to $\boldsymbol{X}^{1:T}$ as $\boldsymbol{X}$, $\boldsymbol{y}^{1:T}$ as $\boldsymbol{y}$, and $\boldsymbol{\tau}^{1:T}$ as $\boldsymbol{\tau}$ in the rest of the paper.
  
\subsection{Related Works}
\textbf{Time Series Anomaly Detection} (TSAD) has advanced from traditional statistical methods~\citep{mclachlan1988mixture,scholkopf1999support,breunig2000lof,tax2004support} to complex Deep Learning (DL) methods~\citep{schmidl2022anomaly,darban2024deeplearningtimeseries}. While DL methods such as forecasting-~\citep{hundman2018detecting,shen2020timeseries} and reconstruction-based models~\citep{su2019robust,audibert2020usad,xu2022anomaly} offer improved detection, they can struggle with rapidly changing data and noisy labels~\citep{golestani2014can,zhou2023detecting,chen2024cluster}. 
Density-based methods~\citep{dai2022graphaugmented,zhou2023detecting}
provide robust performance across scenarios. Recently, large-scale pre-trained models such as AnomalyLLM~\citep{liu2024large2} and AnomalyBERT~\citep{jeong2023anomalybert} have emerged. Yet, most methods focus on statistical dependencies, often overlooking the underlying generation process behind anomalies.

\textbf{Causal Inference} (CI)~\citep{pearl2000models, pearl2016causal} seeks to investigate causal relationships between variables, ensuring robust learning and inference. Integrating DL techniques with CI has shown great promise in recent years, especially in computer vision \citep{zhang2020causal,lv2022causality}, natural language processing~\citep{roberts2020adjusting,tian2022debiasing}, and spatio-temporal data mining~\citep{xia2024deciphering,wang2024nuwadynamics}. In the realm of sequential data, CI is often leveraged to address temporal OoD issues by learning disentangled seasonal-trend~\citep{cost} or environment-specific representations~\citep{yang2022towards} to enhance forecasting accuracy. Though promising, the intrinsic causal mechanisms in TSAD differ from the prediction problem, and the application of CI in this domain remains in its early stages.

\textbf{Normalizing Flows} (NFs)~\citep{tabak2013family,papamakarios2021normalizing} is a powerful technique for density estimation, widely applied in tasks such as image generation~\citep{papamakarios2017masked}. Advanced variants have been developed to enhance models' applicability, e.g.,  RealNVP~\citep{dinh2017density}. Recently, NFs have been explored for anomaly detection across many domains, relying on the assumption that anomalies reside in low-density regions~\citep{rudolph2021same,gudovskiy2022cflow}. 
In the time series realm, following an initial application of NFs for time series forecasting~\citep{rasul2021multivariate}, NFs-based TSAD has been explored by GANF~\citep{dai2022graphaugmented} and MTGFlow~\citep{zhou2023detecting}. Yet, these methods fail to account for the multiple periodicities inherent in time series and overlook the generative processes driving anomalies.

\vspace{-0.5em}
\section{A Causal View on TSAD}\label{sec:causal_view}
 \vspace{-0.5em}
\subsection{Causal Perspective: Generation of Anomalies}\label{sec:scm}
 \vspace{-0.5em}

\begin{wrapfigure}{R}{0.5\textwidth}
\vspace{-2em}
  \includegraphics[width=\linewidth]{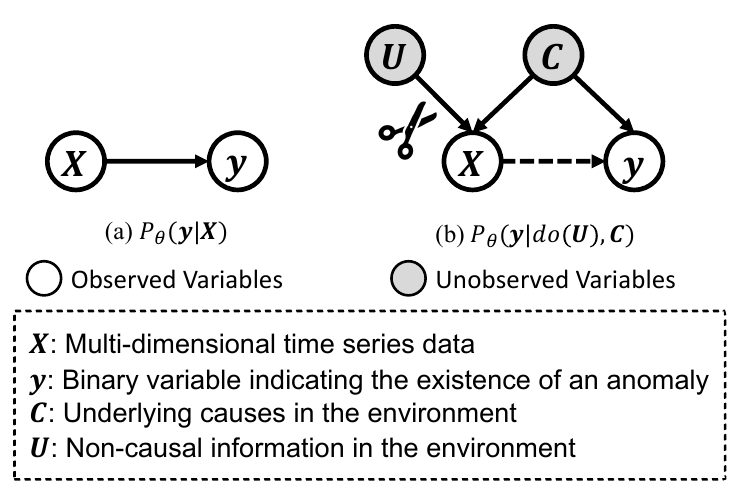}
  \vspace{-1.7em}
  \caption{SCMs of (a) Existing TSAD methods; (b) Time series anomaly generation under real-world scenarios. Solid arrow: causal relationships. Dash arrow: statistical dependencies. Scissors: causal intervention $do(\cdot)$.}\label{fig:scm}
  \vspace{-1.3em}
\end{wrapfigure}

Existing TSAD methods typically infer anomalies $\boldsymbol{y}$ based solely on the input $\boldsymbol{X}$ (Figure~\ref{fig:scm}a), i.e., modeling $P_{\theta}(\boldsymbol{y}|\boldsymbol{X})$, where $P_{\theta}(\cdot)$ denotes the distribution induced by a model $f_{\theta}$. However, real-world scenarios are often more complex than modeling these statistical dependencies between the input and the label, since there exist various underlying factors directly or indirectly influencing the anomaly generation. To address this, we adopt a causal look and introduce an SCM~\citep{pearl2000models} to describe the anomaly generative process and uncover the intrinsic causal relationships between different variables in the context of TSAD, illustrated in Figure\ref{fig:scm}b. Rather than solely modeling $P_{\theta}(\boldsymbol{y}|\boldsymbol{X})$, we propose focusing on $P_{\theta}(\boldsymbol{y}|do(\boldsymbol{U}),\boldsymbol{C})$.

To facilitate understanding, consider an example of cloud computing services. Here, the sequential data $\boldsymbol{X}$ includes the number of slow tasks running on a specific instance within the cloud platform, CPU memory usage, allocated compute resources, and other related metrics that evolve over time. Our goal is to identify issues or anomalies $\boldsymbol{y}$ within this instance caused by true underlying causal factors $\boldsymbol{C}$ from the environment.
Here $\boldsymbol{C} =\left\{\boldsymbol{c}_1, \boldsymbol{c}_2, \cdots, \boldsymbol{c}_N\right\} \in \mathbb{R}^{N \times D_c}$ refers to all latent causal factors, such as ``\textit{hardware failures}'' and ``\textit{network latency}''. $N$ and $D_c$ refer to the number and the dimensions of causal factors, respectively.
Yet, there are some non-causal factors $\boldsymbol{U}$ also in the environment, such as ``\textit{user mis-operations}'' or ``\textit{data collection jitter}'', which may affect the readings of $\boldsymbol{X}$ but do not impact the instance itself, thus unrelated to our detection goal $\boldsymbol{y}$. Thus, an ideal detector is expected to root out the influence of $\boldsymbol{U}$ and focus solely on the causal part $\boldsymbol{C}$. {More discussion and another example can be found in Appendix~\ref{app:scm}.}

\vspace{-0.5em}
\subsection{Causal Backing: Independence Requirement}\label{sec:causal_backing}
  \vspace{-0.5em}

Based on the SCM in Figure~\ref{fig:scm}b, our aim is to detect anomalies by identifying their true underlying causes while eliminating the influence of non-causal factors, i.e., modeling $P_{\theta}(\boldsymbol{y}|do(\boldsymbol{U}),\boldsymbol{C})$. The $do(\cdot)$ operator, as defined in do-calculus, signifies an intervention on the variable~\citep{pearl2016causal}. Directly modeling this operator is challenging because it necessitates learning various latent causes $\boldsymbol{C}$ from the raw input $\boldsymbol{X}$~\citep{arjovsky2019invariant}. Inspired by a previous work~\citep{lv2022causality}, we alternatively leverage a couple of widely-used principles from the causal theory to force the representation of causal factors $\boldsymbol{C}$ we learned to satisfy following key properties.

\textbf{Common Cause Principle}~\citep{reichenbach1991direction} posits that for two statistically dependent variables $X$ and $Y$, there exists a variable $C$ that causally influences both, thereby explaining their dependence by rendering them conditionally independent when conditioned on $\boldsymbol{C}$. Accordingly, the SCM depicted in Figure~\ref{fig:scm}b can be formalized as $ \boldsymbol{X} := f(\boldsymbol{C},\boldsymbol{U})$ and $\boldsymbol{y} := h(\boldsymbol{C}) = h(g(\boldsymbol{X})) $, where $ \boldsymbol{C} \perp \boldsymbol{U} $. Here, $ f $, $ h $, and $ g $ denote unknown structural functions that describe how the observed variables $ \boldsymbol{X} $ and $ \boldsymbol{y} $ are generated from the underlying causes $ \boldsymbol{C} $ and the non-causal variable $ \boldsymbol{U} $.  
This leads to our first property for $\boldsymbol{C}$: it should be independent of $\boldsymbol{U}$.
In this way, for any distribution $P(\boldsymbol{X}, \boldsymbol{y})$, given the causal factor $\boldsymbol{C}$, there exists a conditional distribution $P(\boldsymbol{y}|\boldsymbol{C})$ that represents the invariant mechanism triggering the anomaly within time series.

\textbf{Independent Causal Mechanisms} 
~\citep{scholkopf2012causal,peters2017elements} suggest that the conditional distribution of each variable, given its causes, does not influence other causal mechanisms. 
In other words, none of the factorization of $\boldsymbol{C}$ entails information of others~\citep{lv2022causality}. 
Thus it enforces the mutual independence of the causal factors $\boldsymbol{C} = \{\boldsymbol{c}_1, \boldsymbol{c}_2, \dots, \boldsymbol{c}_N\}$, where $N$ is the number of latent causal factors. 

Therefore, instead of directly learning the causal factors $\boldsymbol{C}$, we enforce them to satisfy the following requirements: \textbf{R1}) $\boldsymbol{C}$ should be independent of $\boldsymbol{U}$, i.e., $\boldsymbol{C} \independent \boldsymbol{U}$, and \textbf{R2}) the components of $\boldsymbol{C}$ should be mutually independent, i.e., $\boldsymbol{c}_1 \independent \boldsymbol{c}_2 \independent \dots \independent \boldsymbol{c}_N$.

 \vspace{-0.5em}
\section{Model Instantiations}\label{sec:method}
\begin{figure*}[!t]
    \centering
    \includegraphics[width=\linewidth]{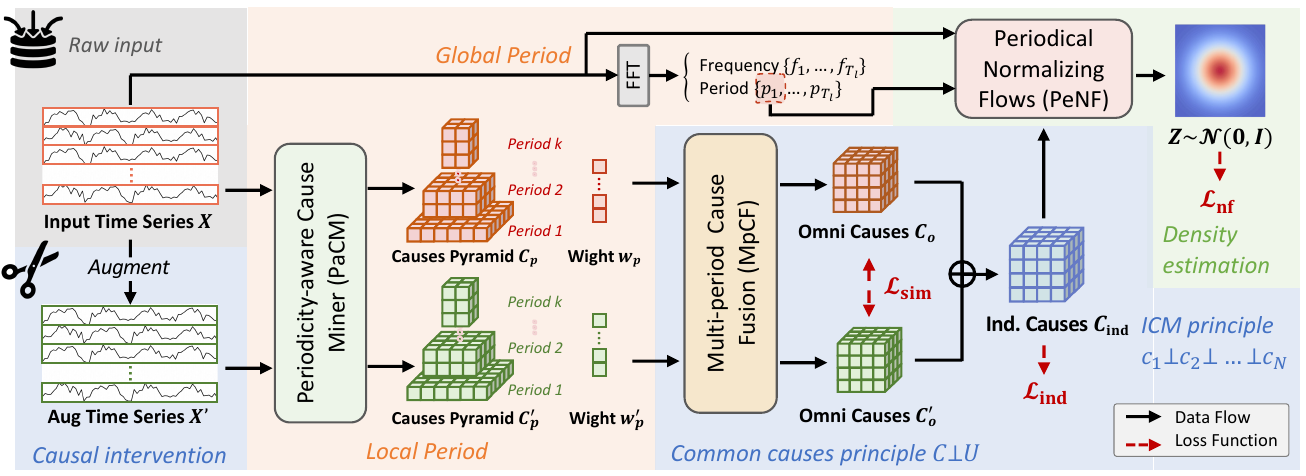}
    \vspace{-0.5em}
    \caption{The pipeline of CaPulse. Different color shaded areas denote solutions for \blueshaded{causal treatments}, \orangeshaded{multiple periodicities}, and \greenshaded{label scarcity \& data imbalance}, respectively. Ind.: Independent. ICM: Independent Causal Mechanisms.}\label{fig:framework}
      \vspace{-1em}
\end{figure*}

 \vspace{-0.5em}
To address the two distinct levels of challenges discussed in the Introduction, we propose a causality-inspired TSAD framework, termed \textbf{CaPulse} (Figure~\ref{fig:framework}). Specifically: 
(1) At the \textbf{mechanistic level}, we incorporate causal treatments (detailed in Section~\ref{sec:causal_treatments}) to satisfy the causal independence requirements \textbf{R1} and \textbf{R2} discussed in the above section.  
(2) At the \textbf{intrinsic data level} (i.e., label scarcity, data imbalance, and multiple periodicities) we introduce a period-aware normalizing flow model to effectively handle these issues (Section~\ref{sec:multi_period}).
It is important to note that the causal perspective in our work serves as a guiding \textit{design principle} for model construction; we do not perform any causal discovery in this study.

\textbf{Framework Overview.} We first provide a brief overview of the framework in Figure~\ref{fig:framework}, followed by a detailed explanation in the subsequent subsections. The input time series $\boldsymbol{X}$ is first augmented to generate $\boldsymbol{X}'$ and subsequently both $\boldsymbol{X}$ and $\boldsymbol{X}'$ are passed through the Periodicity-aware Cause Miner (PaCM) module to obtain $\boldsymbol{C}_{p}$ and $\boldsymbol{C}'_{p}$, i.e., the pyramid representations of latent causal factors at different frequencies. PaCM also outputs the amplitude weights for each frequency, denoted as $\boldsymbol{w}_{p}$ and $\boldsymbol{w}'_{p}$. Next, the Multi-period Cause Fusion (MpCF) module fuses information across different periods based on the amplitude weights and a plugged attention mechanism to generate the omni representations $\boldsymbol{C}_{\text{o}}$ and $\boldsymbol{C}'_{\text{o}}$. A similarity loss $\mathcal{L}_{\text{sim}}$ ensures the consistency between these two representations. Then the final representation $\boldsymbol{C}_{\text{ind}}$ is obtained by averaging them. To ensure the independence of the learned causal factors, we impose an orthogonal loss $\mathcal{L}_{\text{ind}}$. 
After that, Periodical Normalizing Flows (PeNF) takes $\boldsymbol{X}$, the global period $p_{g}$ (obtained by Fast Fourier Transform), and $\boldsymbol{C}_{\text{ind}}$ as inputs 
to estimate the density of $\boldsymbol{X}$ by learning a sequence of invertible transformations, mapping $\boldsymbol{X}$ into a simpler distribution $P(\boldsymbol{Z})$, optimized through the loss $\mathcal{L}_{\text{nf}}$.

 \vspace{-0.5em}
\subsection{Causal Treatments}\label{sec:causal_treatments}
 \vspace{-0.5em}
\textbf{Causal Intervention.} Since $\boldsymbol{C}$ should be separated from $\boldsymbol{U}$ (\textbf{R1}), performing an intervention upon $\boldsymbol{U}$ does not make changes to $\boldsymbol{C}$. We thus leverage causal intervention $do(\cdot)$~\citep{pearl2000models}, to mitigate the negative influence of non-causal factors $\boldsymbol{U}$ and extract causal representations $\boldsymbol{C}$ that are unaffected by $\boldsymbol{U}$~\citep{lv2022causality,zhou2023causal}. In real-world scenarios, non-causal elements (e.g., user misreports) often occur randomly, akin to noise typically found in the high-frequency components of time series data~\citep{gao2021robusttad,xia2024dualprism}. Considering this, we conduct causal intervention by adding noise to the less significant part — the high-frequency part — of the input data to simulate real-world disturbances:
{\footnotesize
\begin{equation}\label{eq:aug}
\boldsymbol{X}' = \operatorname{iFFT}(\operatorname{concat}[\operatorname{FFT}(\boldsymbol{X})_{0:k_h}, \operatorname{FFT}(\boldsymbol{X})_{k_h:T}+\boldsymbol{\eta}]),
\end{equation}
}
where $\operatorname{FFT}(\cdot)$ and $\operatorname{iFFT}(\cdot)$ denote the Fast Fourier Transform and its inverse. $\operatorname{FFT}(\cdot)_{i:j}$ denotes the $i$-th to $j$-th components, $k_h$ refers to the high-frequency threshold, and $\boldsymbol{\eta} \sim \mathcal{N}(0, \sigma^2)$ is the added noise. 
Then we obtain the causal representations $\boldsymbol{C}_o$ and $\boldsymbol{C}'_o \in \real^{{N \times D_c}}$ of $\mx$ and $\mx'$ via PaCM and MpCF modules (detailed in Section~\ref{sec:multi_period}). To ensure the learned information only contains the invariant causal part, we enforce consistency in them and minimize their difference via a similarity loss $\mathcal{L}_{\text{sim}} = \frac{\langle \boldsymbol{C}_o, \boldsymbol{C}'_o \rangle}{\|\boldsymbol{C}_o\| \|\boldsymbol{C}'_o\|}$.

\textbf{Joint Independence}. After obtaining $\boldsymbol{C}_o$ and $\boldsymbol{C}'_o$, the final causal representation $ \boldsymbol{C}_{\text{ind}}$ is obtained by computing the element-wise mean of the two variables.
To enforce the joint independence requirement (\textbf{R2}),  we apply an orthogonal loss that penalizes deviations from independence, achieved by measuring the squared Frobenius norm of the difference between $\boldsymbol{C}_{\text{ind}}^\top \boldsymbol{C}_{\text{ind}}$ and the identity matrix $ \boldsymbol{I} $: 
$ \mathcal{L}_{\text{ind}} = \left\| \boldsymbol{C}_{\text{ind}}^\top \boldsymbol{C}_{\text{ind}} - \boldsymbol{I} \right\|_{F}^{2}$.
This loss encourages the dimensions of $ \boldsymbol{C}_{\text{ind}} $ do not have mutual information, ensuring their independence.

\begin{figure*}[!t]
    \centering
     \vspace{-0.5em}
    \includegraphics[width=\linewidth]{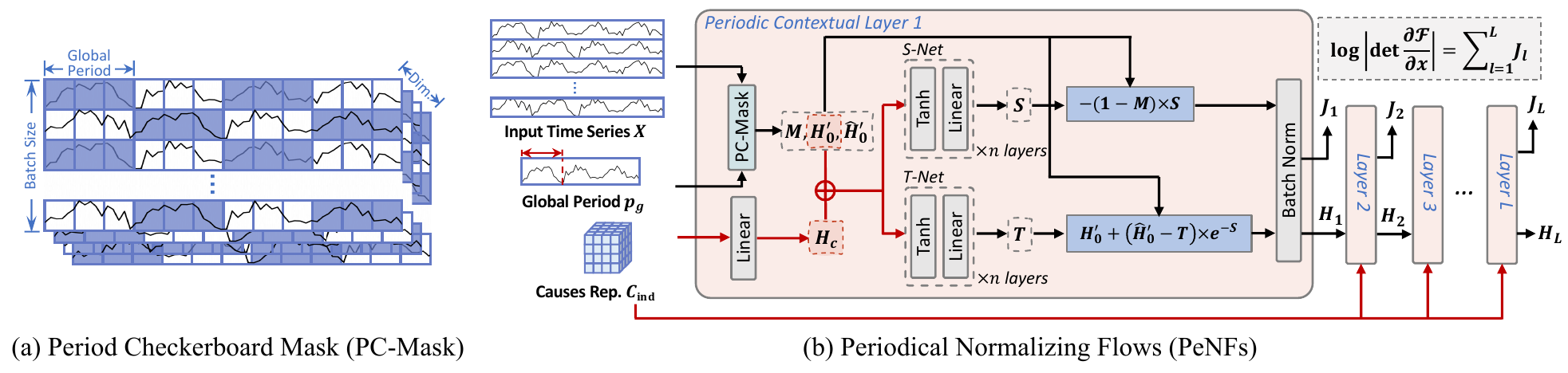}
      \vspace{-0.5em}
    \caption{(a) Masking schemes PC-Mask. (b) Architecture of PeNF, where the black and red arrows represent the data flow for the input and the conditional variable, respectively. }\label{fig:nf}
    \vspace{-1em}
\end{figure*}
 \vspace{-0.5em}
\subsection{Multi-Periodicity Awareness} \label{sec:multi_period}
\vspace{-0.5em}
 
Then we detail the capture of the local and global periodic information (the orange part in Figure~\ref{fig:framework}).

\textbf{Local Periodical Pyramid \& Fusion}. 
We introduce PaCM to extract causal factors for $k$ periodicities, denoted as $ \mathbf{C}_p = \{\boldsymbol{C}_{p1}, \boldsymbol{C}_{p2}, \ldots, \boldsymbol{C}_{pk}\} \in \real^{N \times D_h \times k}$ , along with their corresponding amplitudes $\boldsymbol{w}_p = \{w_{p1}, w_{p2}, \ldots, w_{pk}\} \in \real^{k}$. Inspired by TimesNet~\citep{wu2023timesnet}, this module transforms the input into the frequency domain, selects the top $k$ frequency periods, and reshapes them based on their periodicity. The MpCF module then applies self-attention to compute attention scores $\boldsymbol{a}_p = \{a_{p1}, a_{p2}, \ldots, a_{pk}\} \in \real^{k}$ for each period. After that, it aggregates variables of different periods using both $\boldsymbol{w}_p$ and $\boldsymbol{a}_p$ to generate the final omni representation $\boldsymbol{C}_o \in \real^{N \times D_h}$. The attention mechanism in MpCF dynamically adjusts the importance of each periodic component based on their interactions and dependencies within the time series, rather than relying solely on amplitude when fusing the information across periods. This also enhances the interpretability of the model, discussed in Section~\ref{sec:interpretability}. Due to space constraints, details of PaCM and MpCF are provided in Appendix~\ref{app:detail_lp}.

\textbf{Global Periodical Checkerboard Mask.} To enhance the model's global period awareness, we introduce the PC-Mask scheme tailored to the proposed PeNF (Figure~\ref{fig:nf}). First, for the total length $T_l$ time series with $D$ dimensions $\boldsymbol{X}^{1:T_l}$, we discover the global period $p_{g}$ as follows:
$\boldsymbol{a} = \operatorname{Avg}\big(\operatorname{Amp}\big(\operatorname{FFT}(\boldsymbol{X}^{1:T_l})\big)\big),
    f_{g} = \mathop{\arg\mathrm{max}}\left(\boldsymbol{a}\right),\
    p_{g} = \left\lceil\frac{T_l}{f_{g}}\right\rceil$,
where $\operatorname{Amp}(\cdot)$ and $\operatorname{Avg}(\cdot)$ denotes the average calculation of amplitude values. $\boldsymbol{a}\in\mathbb{R}^{T_l}$ represents the averaged amplitude of each frequency. The $j$-th value $\boldsymbol{a}_{j}$ represents the intensity of the frequency-$j$ periodic basis function, corresponding to the period length $\lceil\frac{T_l}{j}\rceil$. We select the largest amplitude values to obtain the most significant frequencies $f_{g}$, and then we regard its corresponding period length $p_{g}$ as our global period. Next, we use $p_{g}$ to create PC-Mask $\boldsymbol{M} \in \mathbb{R}^{T\times D}$ by a repeating pattern of $ p_g $ zeros followed by $ p_g $ ones (Figure~\ref{fig:nf}a). This process is formulated as $m^{i}_{j} = \left( \left\lfloor \frac{j}{p_g} \right\rfloor \mod 2 \right)$,
where $m^{i}_{j}$ is the element of the mask $\boldsymbol{M}$ at position $(i, j)$, $\lfloor \cdot \rfloor$ denotes the floor function and $\mod$ denotes the modulo operation. This mask will be used for periodicity-awareness, detailed in the following section.

 \vspace{-0.5em}
\subsection{Density Estimation}\label{sec:density_estimation}
 \vspace{-0.5em}

To address the issue of limited labels and imbalanced data, we leverage NFs to achieve an unsupervised density-based anomaly detector. Building on the success of conditioned NFs for time series~\citep{rasul2021multivariate}, we propose PeNF (Figure~\ref{fig:nf}b) with the periodically-awareness introduced by PC-Mask. Overall, PeNF performs the density estimation of the input $\boldsymbol{X}$ conditioned on the causal representation $\boldsymbol{C}_{\text{ind}}$ by learning a sequence of invertible functions $\mathcal{F}$ mapping $\boldsymbol{X}$ into a simple distribution $P(\boldsymbol{Z})$. With the flows parameterized with $\theta$, i.e., $\mathcal{F}_{\theta}:\mathbb{R}^{D}\times\mathbb{R}^{D_h}\to\mathbb{R}^{D}$, where $D_h$ denotes the hidden dimension, the conditioned distribution of $\boldsymbol{X}$ can be expressed as:
{\footnotesize
    \begin{align}\label{eq:px}
{P_{\mathcal{X}}(\boldsymbol{X}|\boldsymbol{C}_{\text{ind}}) }
         {= P_{\mathcal{Z}} (\boldsymbol{Z}|\boldsymbol{C}_{\text{ind}}) \left|\det\frac{\partial{\boldsymbol{Z}}}{\partial{\boldsymbol{X}}} \right|}  {= P_{\mathcal{Z}}(\mathcal{F}_\theta(\boldsymbol{X}, \boldsymbol{C}_{\text{ind}}))\left       |\det\frac{\partial{\mathcal{F}_\theta(\boldsymbol{X}, \boldsymbol{C}_{\text{ind}})}}{\partial{\boldsymbol{X}}} \right|},
\end{align}
}

where $\left|\det(\partial{\mathcal{F}_\theta}/\partial{\boldsymbol{X}}) \right|$ is the Jacobian of $\mathcal{F}_\theta$ at $\boldsymbol{X}$ and $ P_{\mathcal{Z}}$ is the distribution of $\boldsymbol{Z}  \in \mathbb{R}^{T \times D}$ which is chosen to be the standard normal  $ \boldsymbol{z} \sim \mathcal{N}(0, \boldsymbol{I})  \in \mathbb{R}$ in this work. In practice, PeNF takes the PC-Mask $\boldsymbol{M}$ (or the global period $p_g$), the causal representation $\boldsymbol{C}_{\text{ind}}$ and the input data $\boldsymbol{X}$ as its input. Inspired by ~\cite{dinh2017density} and ~\cite{rasul2021multivariate}, we design \textit{periodic contextual layers} to enable NFs aware of periodicity and PeNF consists of $L$ periodic contextual layers, detailed in Appendix~\ref{app:penf}.

 \vspace{-0.5em}
\subsection{Optimization \& Anomaly Measurement} 
 \vspace{-0.5em}
We minimize the total loss: $\mathcal{L} = \mathcal{L}_{\text{nf}} + \alpha \mathcal{L}_{\text{sim}} + \beta\mathcal{L}_{\text{ind}}$, where $\alpha$ and $\beta$ regulate the trade-off of the causal intervention and cause independent loss, and $\mathcal{L}_{\text{nf}}$ is used to optimize the density estimation of {$\boldsymbol{X}$ conditioned on $\boldsymbol{C}_{\text{ind}}$}, denoted as the negative logarithms of the likelihoods in Eq.~\ref{eq:px}:
{\footnotesize
\begin{equation}
\mathcal{L}_{\text{nf}}  = - \sum_{t=1}^T \Big[\log P_{\mathcal{Z}}(\mathcal{F}_\theta(\boldsymbol{x}_t,\boldsymbol{c}_{t})) + \log \left| \det\frac{\partial{\mathcal{F}_\theta(\boldsymbol{x}_t,\boldsymbol{c}_t})}{\partial{\boldsymbol{x}_t}}\right|\Big].
\label{eq:loss_nf}
\end{equation}
}

Density-based approaches act as anomaly detectors based on the widely accepted hypothesis that abnormal instances exhibit lower densities compared to normal ones~\citep{wang2020deep,zhou2024label}. Following prior works~\citep{dai2022graphaugmented,zhou2023detecting}, we compute the anomaly score $\boldsymbol{\tau}$ as the negative logarithm of the likelihood of the input time series $\boldsymbol{X}$ in Eq.~\ref{eq:px}:
{\footnotesize
\begin{align}\label{eq:log_p}
    \boldsymbol{\tau}(\boldsymbol{X}) = - \log P_{\mathcal{X}}(\boldsymbol{X}|\boldsymbol{C}_{\text{ind}}) 
    =  - (\log P_{\mathcal{Z}}(\mathcal{F}_\theta(\boldsymbol{X},\boldsymbol{C}_{\text{ind}})) +  \log \left|\det\frac{\partial{\mathcal{F}_\theta(\boldsymbol{X},\boldsymbol{C}_{\text{ind}})}}{\partial{\boldsymbol{X}}}\right|). 
\end{align}
}

\section{Experiments}\label{sec:experiments}
\vspace{-0.5em}
\subsection{Datasets \& Baselines}
\vspace{-0.5em}
We evaluate CaPulse on seven real-world datasets from different domains, including five commonly used public datasets for TSAD - MSL~\citep{hundman2018detecting}, SMD~\citep{su2019robust}, PSM~\citep{abdulaal2021practical}, WADI~\citep{ahmed2017wadi} - and three cloud services datasets from Alibaba Group
, i.e., Cloud-B, Cloud-S, and Cloud-Y. For comparison, we select eleven TSAD baselines, including 
DeepSVDD~\citep{ruff2018deep}, DeepSAD~\citep{ruff2019deep}, ALOCC~\citep{sabokrou2020deep}, DROCC~\citep{goyal2020drocc}, USAD~\citep{audibert2020usad}, DAGMM~\citep{zong2018deep}, AnomalyTransformer~\citep{xu2022anomaly}, {TimesNet~\citep{wu2023timesnet} and DualTF~\citep{nam2024breaking}}, GANF~\citep{dai2022graphaugmented} and MTGFlow~\citep{zhou2023detecting}. The details of implementation, datasets and baselines are shown in Appendix~\ref{app:CaPulse_setting}, ~\ref{app:dataset} and ~\ref{app:baseline}, respectively.

\begin{table*}[!t]
  \vspace{-1em}
  \centering
  \footnotesize
  \tabcolsep=1.4mm
    \caption{Comparison of 5-run AUROC, presented as the mean values with the corresponding standard deviation. The \bestcolortext{\textbf{best}}/\secondcolortext{\underline{second-best}} results are highlighted. Significance levels ($p < 0.05$) are marked with * (Wilcoxon signed-rank test~\citep{conover1999practical}). See Appendix~\ref{app:significance_analysis} for detailed $p$-values. Ano.Trans.: AnomalyTransformer.}\label{tab:results}
    
\begin{tabular}{l|ccccccc}
\shline
                  & \textbf{Cloud-B}                      & \textbf{Cloud-S}                      & \textbf{Cloud-Y}                        & \textbf{WADI}                         & \textbf{PSM}                          & \textbf{SMD}                          & \textbf{MSL}                          \\\hline

\textbf{DeepSVDD}$^{**}$ & 0.891\tiny{$\pm$0.006} & 0.637\tiny{$\pm$0.085}                         & 0.483\tiny{$\pm$0.064}                         & 0.742\tiny{$\pm$0.013}                         & 0.640\tiny{$\pm$0.069}                         & 0.805\tiny{$\pm$0.048}                         & \secondcolor \underline{0.571}\tiny{$\pm$0.028} \\
\textbf{ALOCC}$^{**}$    & 0.725\tiny{$\pm$0.120}                         & 0.716\tiny{$\pm$0.120}                         & 0.587\tiny{$\pm$0.030}                         & 0.709 \tiny{$\pm$0.080}                         & 0.651\tiny{$\pm$0.120}                         & 0.712\tiny{$\pm$0.060}                         & 0.504\tiny{$\pm$0.016}                         \\
\textbf{DROCC}$^{*}$    & 0.807 \tiny{$\pm$0.080}                         & 0.732\tiny{$\pm$0.06}                         & 0.664\tiny{$\pm$0.110}                         & 0.710\tiny{$\pm$0.090}                         & 0.711\tiny{$\pm$0.180}                         & 0.704\tiny{$\pm$0.080}                         & 0.529\tiny{$\pm$0.069}                         \\
\textbf{DeepSAD}$^{**}$  & 0.867\tiny{$\pm$0.027}                         & 0.642\tiny{$\pm$0.079}                         & 0.453\tiny{$\pm$0.056}                         & 0.723\tiny{$\pm$0.009}                         & 0.644\tiny{$\pm$0.076}                         & 0.818\tiny{$\pm$0.055}                         & 0.521\tiny{$\pm$0.011}                         \\
\textbf{DAGMM}$^{**}$    & 0.775\tiny{$\pm$0.040}                         & 0.707\tiny{$\pm$0.020}                         & 0.660\tiny{$\pm$0.080}                         & 0.749\tiny{$\pm$0.050}                         & 0.633\tiny{$\pm$0.129}                         & {0.837}\tiny{$\pm$0.030}                         & 0.516\tiny{$\pm$0.024}                         \\
\textbf{USAD}$^{**}$     & 0.844\tiny{$\pm$0.076}                         & 0.532\tiny{$\pm$0.090}                         & 0.506\tiny{$\pm$0.056}                         & 0.781\tiny{$\pm$0.030}                         & 0.704\tiny{$\pm$0.019}                         & 0.782\tiny{$\pm$0.023}                         & 0.562\tiny{$\pm$0.001}                         \\
\textbf{Ano.Trans.}$^{*}$     & 0.871\tiny{$\pm$0.009}                         & 0.783\tiny{$\pm$0.048}                         &  0.672\tiny{$\pm$0.082} & 0.763\tiny{$\pm$0.006} & 0.708\tiny{$\pm$0.043} & 0.835\tiny{$\pm$0.054}                         & 0.564 \tiny{$\pm$0.003}                         \\

{\textbf{TimesNet}}     & \secondcolor \underline{0.893}\tiny{$\pm$0.009} & 0.836\tiny{$\pm$0.006}     &  0.727\tiny{$\pm$0.016} & 0.756\tiny{$\pm$0.013} & \secondcolor \underline{0.743}\tiny{$\pm$0.029} & \secondcolor \underline{0.882}\tiny{$\pm$0.010}                         & 0.562 \tiny{$\pm$0.002}                         \\

{\textbf{DualTF}}     & 0.708\tiny{$\pm$0.116}                         & 0.706\tiny{$\pm$0.141}                         &  0.677\tiny{$\pm$0.111} & 0.796\tiny{$\pm$0.030} & 0.727\tiny{$\pm$0.071} & 0.796\tiny{$\pm$0.101}                         & 0.565 \tiny{$\pm$0.003}                         \\\hline

\textbf{GANF}     & 0.857\tiny{$\pm$0.024}                         & 0.805\tiny{$\pm$0.038}                         & \bestcolor \textbf{0.743}\tiny{$\pm$0.056} & \bestcolor \textbf{0.843}\tiny{$\pm$0.005} & {0.725}\tiny{$\pm$0.010} & 0.772\tiny{$\pm$0.055}                         & 0.443\tiny{$\pm$0.037}                         \\
\textbf{MTGFLOW}  & 0.884\tiny{$\pm$0.013}                         & \secondcolor \underline{0.842}\tiny{$\pm$0.028} & 0.728\tiny{$\pm$0.044}                         & 0.822\tiny{$\pm$0.018}                         & 0.721\tiny{$\pm$0.035}                         &  0.836\tiny{$\pm$0.023} & 0.570\tiny{$\pm$0.003}                         \\
\textbf{CaPulse} (Ours)    & \bestcolor \textbf{0.926}\tiny{$\pm$0.007} & \bestcolor \textbf{0.887}\tiny{$\pm$0.021} & \secondcolor \underline{0.741}\tiny{$\pm$0.030} & \secondcolor \underline{0.830}\tiny{$\pm$0.029} & \bestcolor \textbf{0.753}\tiny{$\pm$0.042} & \bestcolor \textbf{0.901}\tiny{$\pm$0.009} & \bestcolor \textbf{0.604}\tiny{$\pm$0.017}\\\shline                    
\end{tabular}
\end{table*}

 \vspace{-0.5em}
\subsection{Empirical Results}\label{sec:exp_results}
\begin{figure*}[!t]
    \centering
    \vspace{-1.5em}
    \includegraphics[width=\linewidth]{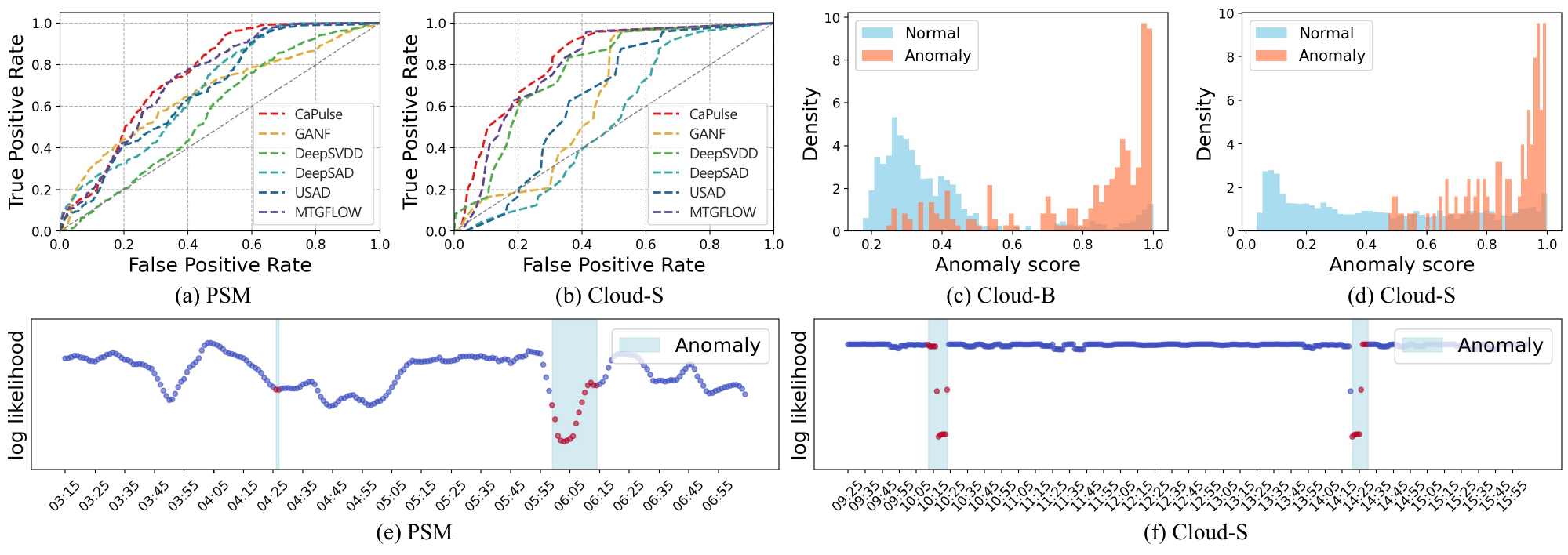}
       \vspace{-1.5em}
    \caption{(a) and (b) are comparisons of AUROC curves for various models on the PSM and Cloud-S datasets, respectively. (c) and (d) are the density plots of anomaly scores for normal and anomalous instances in the Cloud-B and Cloud-S datasets. (e) and (f) visualize the log-likelihood in PSM and Cloud-S datasets.}\label{fig:performance}
      \vspace{-1em}
\end{figure*}

 \vspace{-0.5em}
\textbf{Model Comparison.} We follow previous density-based methods~\citep{dai2022graphaugmented,xu2023density} to evaluate models using the Area Under the Receiver Operating Characteristic (AUROC), where higher values indicate better performance. \textit{Quantitatively}, Table~\ref{tab:results} reports the mean and standard deviation (STD) of AUROC scores over 5-run experiments. From these results, we can observe: 1) CaPulse achieves the highest AUROC on five out of seven datasets and ranks second on the remaining two, highlighting its robustness and consistency across various datasets. 2) CaPulse exhibits low variance, reflected by its small STD values, outperforming most baselines and demonstrating its generalizability. 3) While other NFs-based models (MTGFlow and GANF) perform well on specific datasets, they generally fall short of CaPulse, especially in cloud systems where the underlying causality of anomaly is crucial. \textit{Graphically}, Figure~\ref{fig:performance}a and ~\ref{fig:performance}b present the AUROC curves for two datasets, which illustrate the trade-off between the True Positive Rate (TPR) and False Positive Rate (FPR) across different threshold settings. The results show that CaPulse outperforms the baseline models by achieving higher TPRs at lower FPRs. 

\textbf{Anomaly Score Distributions.} We first provide anomaly score distributions of the proposed model on two datasets in Figures~\ref{fig:performance}c and~\ref{fig:performance}d. Blue bars represent normal data, while red bars indicate anomalies. Anomalies cluster toward the higher end of the score range, near 1. For Cloud-B, normal points are spread between 0.2 and 0.6, while anomalies concentrate around 0.9 and above. In Cloud-S, the separation is more pronounced, with most anomalies scoring above 0.8, demonstrating the model's ability to effectively distinguish between normal and anomalous data. 

\textbf{Log-Likelihood.} The log-likelihood behavior during anomalies of two datasets are shown in Figure~\ref{fig:performance}e and~\ref{fig:performance}f, respectively, where the shaded areas represent true anomalies. According to the figures, in PSM, log-likelihood drops sharply at the anomaly around 06:05, indicating the model's lower confidence during abnormal events. Similarly, in Cloud-S, the log-likelihood decreases significantly at around 10:15 and 14:25, correctly aligning with the true anomaly. These results confirm the model's effectiveness in detecting anomalies by observing clear drops in likelihood during anomalous intervals. 

 \vspace{-0.5em}
\subsection{Interpretability Analysis}\label{sec:interpretability}
 \vspace{-0.5em}
\begin{wrapfigure}{R}{0.5\textwidth}
\vspace{-2em}
  \includegraphics[width=\linewidth]{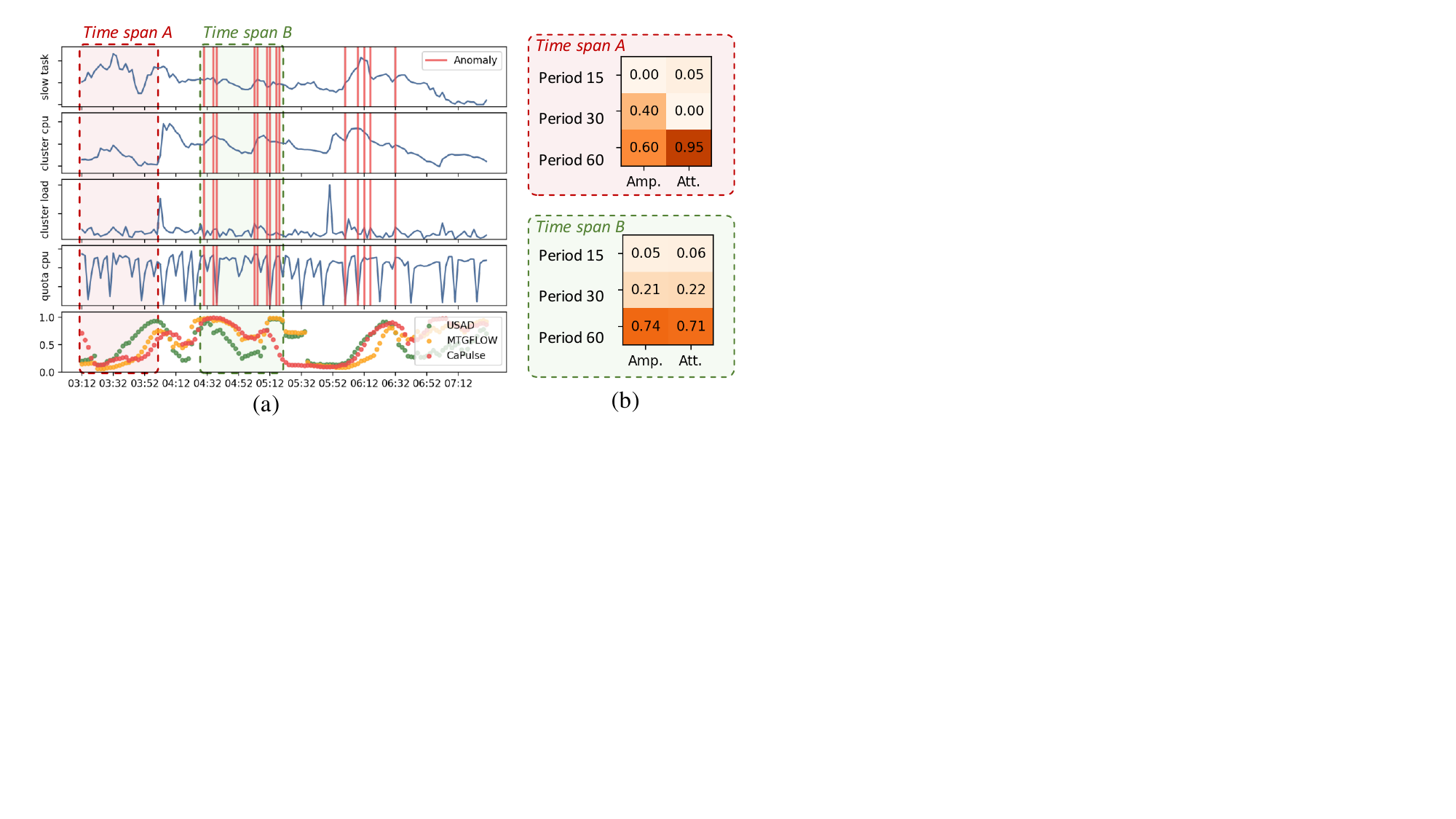}
  \vspace{-2em}
    \caption{(a) Time series data with ground truth anomaly (\textit{first four rows}) and predicted anomaly scores from CaPulse and other methods (\textit{last row}). (b) Corresponding weights for cause pyramids for Time spans A and B. Amp.: Amplitude. Att.: Attention.}\label{fig:interpretability}
     \vspace{-1em}
\end{wrapfigure}

\textbf{True Causal Factor Identification.} Figure~\ref{fig:interpretability}a presents the time series data, ground truth anomalies, and anomaly scores predicted by CaPulse, USAD, and MTGFlows on the Cloud-S dataset. The first four rows show different metrics changing over time and the red lines represent the anomaly labels. Time span A is a period of normal operation, while Time span B highlights abnormal events occurring in the instance (i.e., virtual machine) in the cloud computing platform. In Time span A, while there is a rise in slow tasks at around 03:52, other metrics such as CPU usage and system load remain stable, suggesting \textit{user misoperation} might be a possible cause for it rather than a true anomaly. CaPulse captures these underlying causal factors, demonstrating its ability to detect non-obvious anomalies, while USAD does not and assigns a higher anomaly score. In contrast, during Time span B, subtle anomalies occur despite no visible abrupt changes. CaPulse captures these underlying causal factors, demonstrating its robustness in detecting non-obvious anomalies.  Although USAD and MTGFlows also recognize this anomaly, they continue assigning high scores for 20 minutes after Time span B, failing to recognize the return to normal operation.

\textbf{Significance of Attention Mechanism.} The elevated anomaly scores predicted by CaPulse (bottom row) during Time span B align with the ground truth. Figure~\ref{fig:interpretability}b further illustrates how feature weights differ between the two time spans. 
When fusing causal factors across different periods, amplitude weights alone cannot effectively prioritize critical periods for identifying anomalies, whereas attention scores provide this capability. As shown in Figure~\ref{fig:interpretability}b, during Time span A, although the amplitude weights assign similar importance to Periods 30 and 60, the high attention score for Period 60 (0.95) highlights that long-term features are more relevant for capturing causal factors. This is particularly important when addressing short-term ``user misoperations'', where focusing only on short-term patterns could result in misinterpretations. The attention mechanism mitigates this risk by directing focus to the most relevant periods. 

\begin{figure*}[!t]
    \centering
    \includegraphics[width=0.85\linewidth]{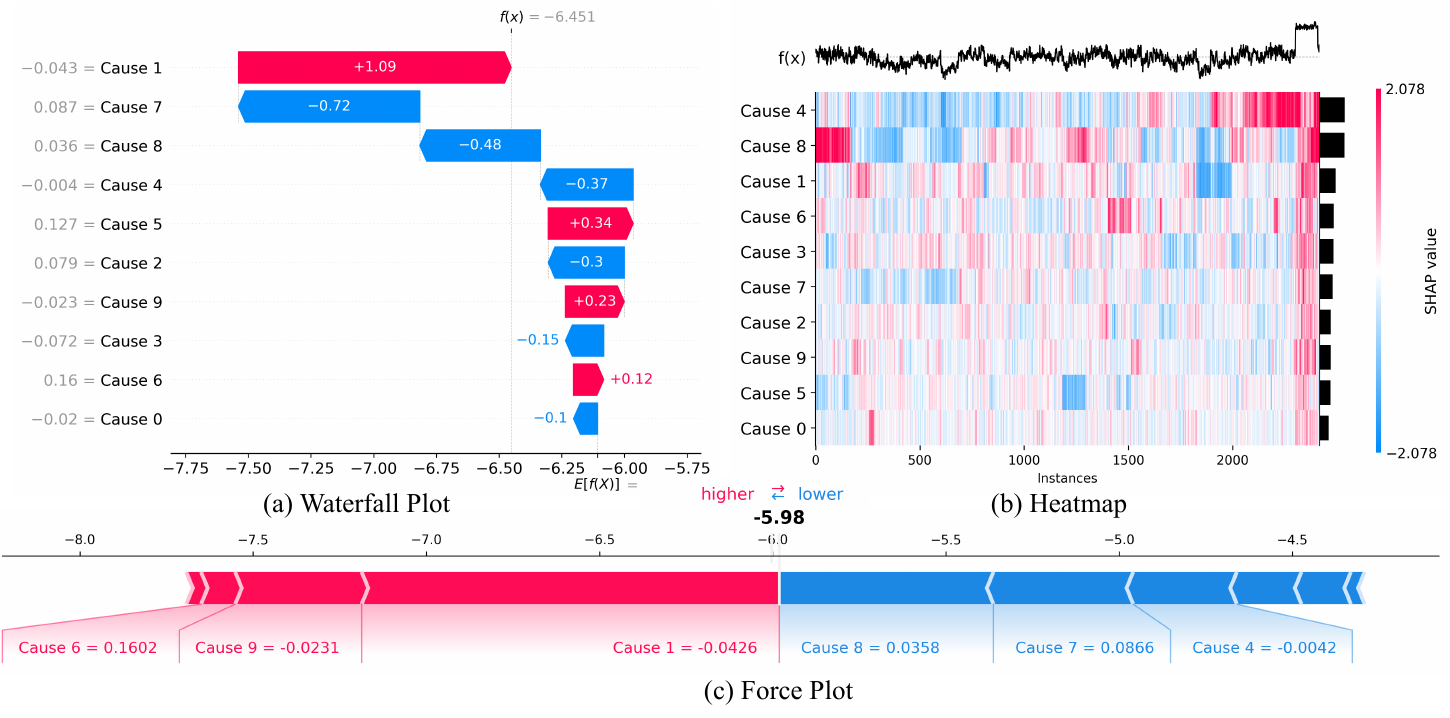}
      \vspace{-1em}
    \caption{(a) Waterfall plot: SHAP values for an individual prediction showing contributions from each cause. (b) Heatmap: SHAP values across multiple instances and causes. (c) Force plot: individual feature contributions for a specific instance. Cause $i$: the $i$-th latent causal factor $\boldsymbol{c}_{i}$.}\label{fig:interpretability_shap}
     \vspace{-1em}
\end{figure*}

\textbf{Interpretability of Causal Representations.} Next, we analyze the interpretability of the ``causal rhythm'' learned by CaPulse, i.e., the representation of latent causal factors $\boldsymbol{C}_{\text{ind}}$. The analysis uses the Cloud-S dataset, {with 10 latent causal factors ($N=10$)} denoted as $\{\boldsymbol{c}_1, \boldsymbol{c}_2, \dots, \boldsymbol{c}_{10}\}$. We then present an interpretability analysis using the SHapley Additive exPlanations (SHAP)~\citep{NIPS2017_7062}, which helps explain how each latent cause contributes to the anomaly detection. The interpretability results are visualized in Figure~\ref{fig:interpretability_shap}. {Red (positive SHAP values) indicates a push towards anomaly detection, while blue (negative SHAP values) indicates a shift towards normal behavior.} According to the result, we have the following observations: 
1) The waterfall plot in Figure~\ref{fig:interpretability_shap}a presents the contribution of each cause for a specific sample, where $\boldsymbol{c}_1$ contributes the most positively, {pushing the prediction towards the anomaly exist}, yet $\boldsymbol{c}_7$ has the most significant negative impact, {shifting the prediction towards normal behavior}. 
2) The heatmap in Figure~\ref{fig:interpretability_shap}b provides a global overview of how the causes impact identifying anomalies across multiple samples. {Each row represents a latent cause, and each column represents a sample. $\boldsymbol{c}_1$, $\boldsymbol{c}_4$ and $\boldsymbol{c}_8$ show consistently high positive SHAP values for many instances, while $\boldsymbol{c}_5$ and $\boldsymbol{c}_7$ stand out with significant negative SHAP values across many instances.}
3) The force plot in Figure \ref{fig:interpretability_shap}c provides a detailed view of how these causes push or pull a specific detection from the average value to the final prediction. {In this sample, $\boldsymbol{c}_6$ drives the prediction towards anomaly, while $\boldsymbol{c}_9$ highly recognizes the sample is normal. $\boldsymbol{c}_1$ and $\boldsymbol{c}_7$ show moderate contributions.}

In summary, causes like $\boldsymbol{c}_1$ how consistently demonstrate a strong positive influence on anomaly detection, indicating that its representation is closely linked to anomaly-indicating patterns {(e.g., "hardware failure" in a cloud service context)}. Conversely, causes like $\boldsymbol{c}_7$ tend to shift predictions toward normal behavior, suggesting that these causes are more reflective of regular instances {(e.g., ``users' misperception'')}. 
Detailed experimental settings and plot explanations are provided in Appendix~\ref{app:shap}.

 \vspace{-0.5em}
\subsection{Ablation Study \& Hyperparameter Sensitivity}\label{sec:ablation}
\vspace{-0.5em}

\begin{wraptable}{R}{0.5\textwidth}
    \centering
    \small
    \tabcolsep=2.9mm
    \vspace{-2em}
    \caption{Variant results on two datasets.}\label{tab:ablation}
    \begin{tabular}{l|ll}
    \shline 
    \textbf{Variant}  & \textbf{SMD}                               & \textbf{Cloud-S}                         \\
    \hline
    \textbf{w/o CI}   &  \secondcolor \underline{0.890}\tiny{$\pm$0.015} (\blue{$\downarrow$1.87\%}) & 0.825\tiny{$\pm$0.056} (\blue{$\downarrow$6.99\%}) \\
    \textbf{w/o ICM}  & 0.884\tiny{$\pm$0.010} (\blue{$\downarrow$2.54\%}) & 0.848\tiny{$\pm$0.005} (\blue{$\downarrow$4.40\%}) \\
    \textbf{w/o Attn} & 0.888\tiny{$\pm$0.012} (\blue{$\downarrow$2.09\%}) & \secondcolor \underline{0.859}\tiny{$\pm$0.016} (\blue{$\downarrow$3.16\%}) \\
    \textbf{w/o GP}   & 0.889\tiny{$\pm$0.015} (\blue{$\downarrow$1.98\%}) & 0.856\tiny{$\pm$0.011} (\blue{$\downarrow$3.49\%}) \\\hline
    \textbf{CaPulse}    & \bestcolor \textbf{0.901}\tiny{$\pm$0.009} &  \bestcolor \textbf{0.887}\tiny{$\pm$0.021} \\\shline                  
    \end{tabular}
    \vspace{-1em}
\end{wraptable}

\textbf{Effects of Core Components.} To evaluate the contribution of each core component in CaPulse, we conducted an ablation study using the following variants: 
a) \textbf{w/o CI}, which removes causal intervention and the similarity loss;
b) \textbf{w/o ICM}, which excludes the ICM principle, thereby not ensuring joint independence of causal factors;
c) \textbf{w/o Attn}, which omits the attention mechanism used for fusing multi-period features; 
and d) \textbf{w/o GP}, which excludes global period information by not applying the PC-Mask in PeNFs.
Table~\ref{tab:ablation} reports their AUROC results across two datasets, showing that all components contribute significantly to the model's overall performance. Notably, for Cloud-S, excluding causality-related components (\textbf{w/o CI} and \textbf{w/o ICM}) results in a marked performance degradation, underscoring the importance of causal mechanisms in cloud services. 
More ablation results are presented in Appendix~\ref{app:ablation}.

\begin{wrapfigure}{R}{0.5\textwidth}
\vspace{-1em}
  \includegraphics[width=\linewidth]{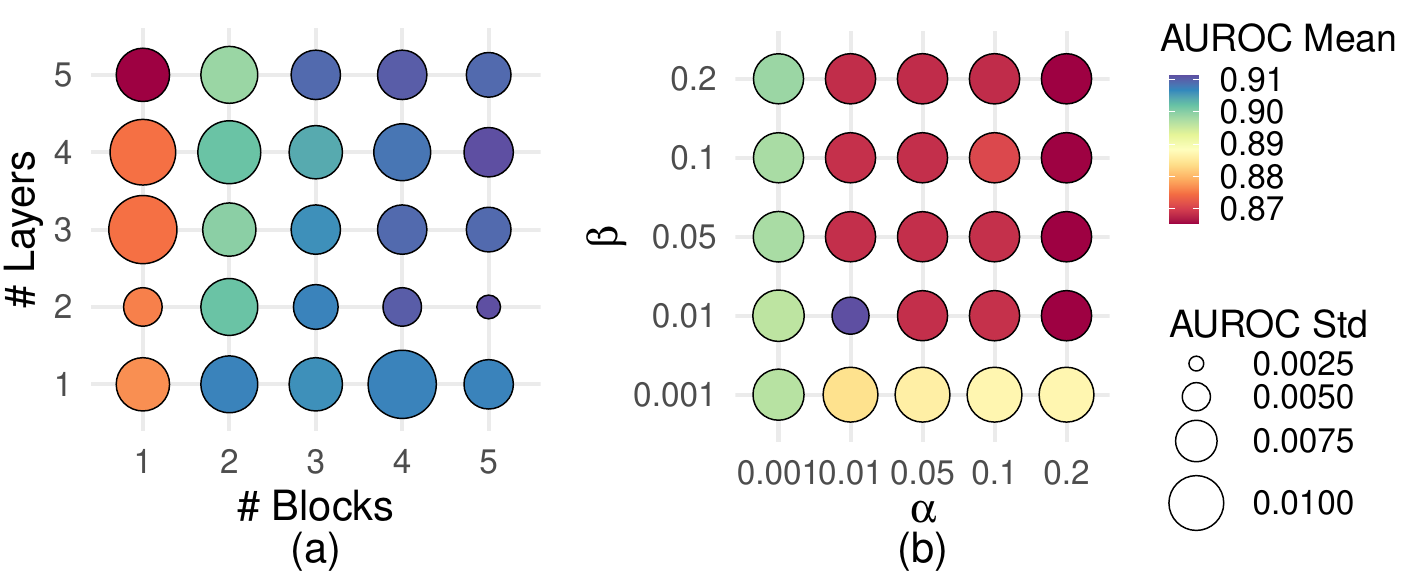}
  \vspace{-1.5em}
    \caption{Study on hyperparameter on AUROC for SMD dataset.}\label{fig:ablation}
    \vspace{-1em}
\end{wrapfigure}

\textbf{Hyperparameter Sensitivity.} Figure~\ref{fig:ablation} illustrates the impact of different configurations of \# Layers, \# Blocks, and the balance coefficients in the loss function, $\alpha$ and $\beta$, on the model's AUROC performance for the SMD dataset. In Figure~\ref{fig:ablation}a, increasing the number of blocks consistently improves performance, while the number of layers has a lesser effect, with the best AUROC achieved at 2 layers and 5 blocks. Figure~\ref{fig:ablation}b reveals the sensitivity to $\alpha$ and $\beta$, showing optimal AUROC when both parameters are set around 0.01. This underscores the need to balance the contributions of different loss terms for optimal performance and stability.

\section{Conclusion}\label{sec:conclusion}
In this paper, we present the first attempt to take a causal intervention-based perspective for TSAD and implement it within a deep learning framework. 
Concretely, building on the proposed SCM, we introduce CaPulse, a causality-driven deep learning model designed to detect anomalies by leveraging causal tools while addressing key challenges in TSAD, including label scarcity, data imbalance, and multiple periodicities. Extensive experiments on seven datasets across five domains demonstrate CaPulse is equipped to effectively detect both subtle and significant deviations, enhancing interpretability and robustness. 


\bibliography{ref}
\bibliographystyle{ACM-Reference-Format}

\newpage
\appendix

\section{Normalizing Flows for Time Series} \label{app:cnf}

\textbf{Normalizing Flows.} Normalizing Flows (NFs)~\citep{tabak2013family,papamakarios2021normalizing} are a powerful technique for density estimation, widely utilized in tasks such as image generation~\citep{papamakarios2017masked}. Essentially, NFs are invertible transformations that map data from an input space $\mathbb{R}^D$ to a latent space $\mathbb{R}^D$, such that a complex distribution $P_\mathcal{X}$ on the input space $\boldsymbol{X} \in \mathbb{R}^{D}$ is transformed into a simpler distribution $P_\mathcal{Z}$ in the latent space $\boldsymbol{Z} \in \mathbb{R}^{D}$. These mappings, $\mathcal{F} \colon \mathcal{X} \mapsto \mathcal{Z}$, are typically constructed as a series of invertible functions. By utilizing the change of variables formula, the probability density function $P_{\mathcal{X}}(\boldsymbol{X})$ is expressed as:

\begin{equation}
P_{\mathcal{X}}(\mathbf{X}) = P_{\mathcal{Z}}(\mathbf{Z}) \left| \det \left( \frac{\partial \mathcal{F}(\mathbf{X})}{\partial \mathbf{X}} \right) \right|,
\end{equation}

where $\frac{\partial \mathcal{F}(\boldsymbol{X})}{\partial \boldsymbol{X}}$ is the Jacobian matrix of the transformation $\mathcal{F}$ at $\boldsymbol{X}$. NFs offer two key advantages: both the inverse transformation $\boldsymbol{X} = \mathcal{F}^{-1}(\boldsymbol{Z})$ and the computation of the Jacobian determinant can be efficiently computed, with the determinant calculation typically taking $O(D)$ time. This enables the following expression for the log-likelihood of the data under the flow:
\begin{equation}\label{logp}
\log P_{\mathcal{X}}(\mathbf{X}) = \log P_{\mathcal{Z}}(\mathbf{Z}) + \log | \det(\partial \mathbf{Z} / \partial \mathbf{X})|.
\end{equation}

\textbf{Temporal Conditioned Normalizing Flows.} To adapt NFs for time series data, temporal conditioned flows introduce additional conditional information, denoted as $\mathbf{h} \in \mathbb{R}^{D_h}$, which may have a different dimension from the input. The flow is now expressed as $\mathcal{F} \colon \mathbb{R}^{D} \times \mathbb{R}^{D_h} \to \mathbb{R}^{D}$, allowing for conditioning on temporal features. The log-likelihood of the time series $\mathbf{X}$, conditioned on the temporal context $\mathbf{h}$, is given by:
\begin{equation}\label{eqn:cnf}
\log P_{\mathcal{X}}(\mathbf{X}|\mathbf{h}) = \log P_{\mathcal{Z}}(\mathcal{F}(\mathbf{X}; \mathbf{h})) + \log |\det (\nabla_{\mathbf{X}} \mathcal{F}(\mathbf{X}; \mathbf{h}))|.
\end{equation}

\textbf{Coupling Layers.} One of the key innovations in NFs proposed by a widely-used variant RealNVP~\citep{dinh2017density} is the use of \emph{coupling layers}, which simplify the computation of the Jacobian determinant. In a coupling layer, part of the input remains unchanged, while another part is transformed. Specifically, the transformation is defined as:
\begin{equation}
\begin{cases}
    \boldsymbol{Y}^{1:d} = \boldsymbol{X}^{1:d}, \\
    \boldsymbol{Y}^{d+1:D} =  \boldsymbol{X}^{d+1:D} \odot \exp(\mathcal{S}_{\theta}(\boldsymbol{X}^{1:d})) + \mathcal{T}_{\theta}(\boldsymbol{X}^{1:d}),
\end{cases}
\end{equation}
where $\odot$ represents element-wise multiplication, $\mathcal{S}(\cdot)$ is a scaling function, and $\mathcal{T}(\cdot)$ is a translation function, both parameterized by $\theta$. The coupling layer thus enables efficient transformations by only modifying part of the input at a time. To achieve complex, nonlinear density mappings, multiple coupling layers are stacked, alternating which dimensions are transformed at each layer. This ensures that all dimensions are transformed over the course of the flow, while keeping computations efficient.

\section{Details of CaPulse}

\subsection{Rationale behind the design of CaPulse}\label{app:resonale}
CaPulse are deliberately co-designed to tackle two distinct levels of challenges TSAD, as outlined in the Introduction:

(1) \textbf{\textit{At the mechanistic level}}, TSAD demands understanding why anomalies occur. We address this by introducing a SCM grounded in the principle of independent mechanisms to guide the model design (Section~\ref{sec:causal_treatments}), enhancing the generalization and the interpretability.

(2) \textbf{\textit{At the data level}}, real-world time series commonly suffer from issues such as label scarcity, data imbalance, and multi-periodicity. To mitigate these, we develop period-aware normalizing flows (Section~\ref{sec:multi_period} and ~\ref{sec:density_estimation}), which perform expressive density estimation and explicitly model periodic structures. This design enables the model to detect rare or subtle anomalies even under limited supervision.

This integrated design ensures that CaPulse delivers interpretable, causally grounded, and fine-grained anomaly scores.

\subsection{Architecture of PaCM \& MpCF}\label{app:detail_lp}

\begin{figure}[h]
    \centering
    \includegraphics[width=0.9\linewidth]{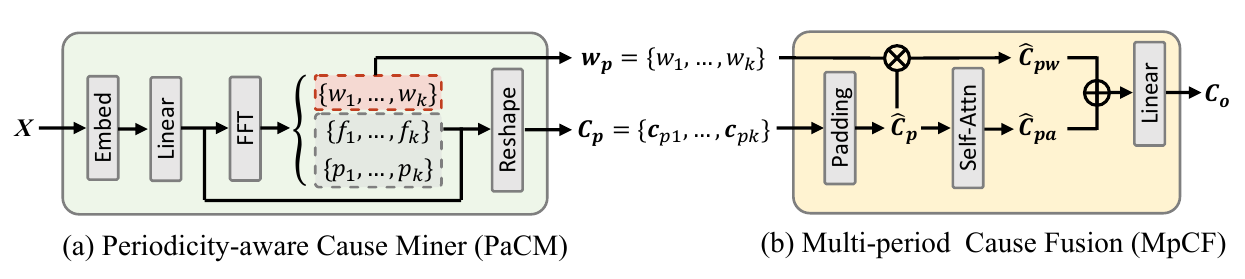}
    \vspace{-1.5em}
    \caption{Architecture of the proposed PaCM and MpCF modules.}\label{fig:local_period}
\end{figure}

We design PaCM and MpCF to handle multiple local periodicities in time series data by extracting and fusing periodic information at various levels, illustrated in Figure~\ref{fig:local_period}a and ~\ref{fig:local_period}b, respectively. Inspired by TimesNet~\citep{wu2023timesnet}, PaCM is responsible for capturing different period levels of information from the input time series $\boldsymbol{X}$. PaCM first embeds the input time series $\boldsymbol{X}$, followed by a linear transformation to project the input into a higher-dimensional space. Next, an FFT is applied to obtain the frequency components $\{f_1, f_2, \dots, f_k\}$ and their corresponding periodicities $\{p_1, p_2, \dots, p_k\}$ and the amplitude weight $\boldsymbol{w}_p = \{w_1, w_2, \dots, w_k\}$. The output of PaCM is a set of reshaped latent causal factors $ \mathbf{C}_p = \{\boldsymbol{C}_{p1}, \boldsymbol{C}_{p2}, \ldots, \boldsymbol{C}_{pk}\} $, where each $\mathbf{C}_{pi}$ represents a representation for the $i$-th period to consist the pyramid $\mathbf{C}_p \in \real^{N \times D_h \times k}$.

MpCF is designed to fuse the multi-period information extracted by PaCM. MpCF begins by padding the causal factors from different periods, followed by applying a self-attention mechanism to compute attention scores for each period. These attention scores, along with the amplitude weights $\boldsymbol{w}_p$, are used to weight the periodic components and produce the final omni-causal representation $\hat{\boldsymbol{C}}_p$. The final output of MpCF is the fused causal representation $\boldsymbol{C}_o \in \real^{N \times D_h}$, which integrates the most relevant information from all periods. The advantages of the attention mechanism are demonstrated by the improvement of performance (see Section~\ref{sec:interpretability}) and interpretability (see Section~\ref{sec:ablation}). Together, PaCM and MpCF effectively capture and fuse periodic information, enabling the model to handle complex, multi-periodic time series data.

\subsection{Details of periodic contextual layer}\label{app:penf}
In Periodical Normalizing Flows (PeNFs), illustrated in Figure~\ref{fig:nf}b, we design the periodic contextual layers to enable NFs aware of periodicity and PeNF consists of $L$ periodic contextual layers. In the $l$-th layer, there are two outputs: $\boldsymbol{H}_{l}$ and $\boldsymbol{J}_l$. The first output will be passed to the next layer for further updates, while the second output will be accumulated across layers and contribute to the final Jacobian variable $\log \left|\det(\partial{\mathcal{F}_\theta}/\partial{\boldsymbol{X}}) \right|$, which be used for optimize and detailed in the next section. 
To obtain $\boldsymbol{H}_{l}$, we use the mask $\boldsymbol{M}$ derived based on the global period $p_g$ to select part of the input $\boldsymbol{H}_{l-1}$ to remain unchanged: $\boldsymbol{H}_{l-1}' = \boldsymbol{H}_{l-1} \odot \boldsymbol{M} $, where $\odot$ denotes the Hadamard product and $\boldsymbol{H}_{0} = \boldsymbol{X}$. The remaining part of the input, $\hat{\boldsymbol{H}}_{l-1}' = \boldsymbol{H}_{l-1} \odot (\boldsymbol{I} - \boldsymbol{M})$, is transformed via functions of the unaltered variables. Thus, in the $l$-th layer, $\boldsymbol{H}_{l}$ will be:
{\footnotesize
\begin{equation}
\boldsymbol{H}_{l} = \boldsymbol{H}_{l-1}' + (\hat{\boldsymbol{H}}'_{l-1} - \mathcal{T}_\theta(\boldsymbol{H}'_{l-1}, \boldsymbol{H}_c)\odot \exp\left( -\mathcal{S}_\theta(\boldsymbol{H}'_{l-1}, \boldsymbol{H}_c \right)) ),
\end{equation}
}
where $\mathcal{S}_\theta(\cdot)$ and $\mathcal{T}_\theta(\cdot)$ are scaling and translation functions parameterized by neural networks with $\theta$, and $\boldsymbol{H}_c \in \real^{T \times D_h}$ is the latent variable obtained by a linear transformation from $\boldsymbol{C}_{\text{ind}}$. Then, a number of these periodic contextual layers mapping are composed together: $\boldsymbol{X} \rightarrow \boldsymbol{H}_1 \rightarrow \boldsymbol{H}_2 \rightarrow \cdots \rightarrow \boldsymbol{H}_{L} \rightarrow \boldsymbol{Z}$.

\subsection{Orthogonality loss for independence}\label{app:orthogonality_loss}

In Capulse, we use orthogonality loss to force joint independence of the representation $\boldsymbol{C}_{\text{ind}}$. Here we explain how orthogonality loss can be used to achieve it. 

If $ X $ and $ Y $ are independent random variables, then their expectations satisfy the relation: $\mathbb{E}[XY] = \mathbb{E}[X]\mathbb{E}[Y]$. Orthogonality is defined as: two objects being orthogonal if their inner product is zero. In the context of random variables, the inner product can be defined when the variables are square-integrable, i.e., they satisfy: $\mathbb{E}[X^2] < \infty \quad \text{and} \quad \mathbb{E}[Y^2] < \infty$, which is satisfied in our context. For such random variables, the inner product is given by: $(X, Y) := \mathbb{E}[XY]$. This definition aligns with the requirements of the Cauchy-Schwarz inequality, which ensures that this inner product satisfies the properties of a valid geometric structure. Now, consider $ X $ and $ Y $ after subtracting their means, so that:
$\mathbb{E}[X] = 0 \quad \text{and} \quad \mathbb{E}[Y] = 0$. By construction: $\mathbb{E}[X]\mathbb{E}[Y] = 0$. If $ X $ and $ Y $ are orthogonal (i.e., $\mathbb{E}[XY] = 0 $), then their inner product vanishes: $\mathbb{E}[XY] = 0$. In this setup, orthogonality implies that the variables guarantee that $ \mathbb{E}[XY] = \mathbb{E}[X]\mathbb{E}[Y] $.

\subsection{Causal representation analysis}\label{app:shap}

In Section~\ref{sec:interpretability}, we analyze the interpretability of the ``causal rhythm'' learned by the proposed model. Here we provide details on the analysis experiment. The experiment was conducted on the Cloud-S dataset, with the number of latent causal factors set to 10, thus resulting in 10 distinct learned causal representations. We present an interpretability analysis using SHAP~\citep{NIPS2017_7062}, SHAP helps explain how each latent cause contributes to the anomalies. Specifically, we first train an XGBoost classifier~\citep{chen2016xgboost} using the learned causal representations to predict the anomaly labels. The SHAP values derived from this model quantify the contribution of each cause to the prediction—indicating how much each cause increases or decreases the likelihood of an anomaly—thereby providing interpretability to the learned representations. For clarity in the analysis, we refer to the latent causes as $\boldsymbol{c}_1$ through $\boldsymbol{c}_{10}$, and the following 'model' is the XGBoost instead of CaPulse. The results are visualized in three SHAP plots (Figure~\ref{fig:interpretability_shap}), each offering unique insights into how individual or grouped causes influence the model's predictions. We have already presented the observation in the main text, so here we just give some explanation about these SHAP plots as follows:

\begin{itemize}[leftmargin=*]
    \item The waterfall plot presents the contribution of each cause for a specific instance (one sample). Starting from the average output of the XGBoost model, the contribution of each cause pushes the prediction either towards predicting an anomaly (in red) or towards predicting normal behavior (in blue). 

    \item The heatmap provides a global overview of how the causes impact predictions across multiple instances. Each row represents a learned cause, and each column represents an instance from the dataset. The color intensity indicates the SHAP value, with red representing a positive contribution towards predicting anomalies and blue representing a negative contribution towards normal behavior. 

    \item The force plot provides a detailed view of how causes push or pull a specific prediction from the base value to the final predicted score. Red arrows represent causes that increase the predicted score (i.e., lead towards an anomaly), while blue arrows represent causes that decrease the score (i.e., lead towards normal behavior). 
\end{itemize}

\subsection{Computational Complexity}\label{app:complexity}
For simplicity, we omit hidden dimensionality in the following analysis. Given that $T$ denotes the number of time steps, the computational complexity of the FFT process is $\mathcal{O}(DT \log T)$, where $D$ refers to the input time series dimension, which is performed in obtaining the global and the local periods. The first stage, i.e., getting the global period is a preprocessing step for the dataset and, thus is not included in the training process. The second stage, i.e., getting the local period occurs within the PaCM module. Additionally, the attention mechanism in the MpCF module introduces a complexity of $\mathcal{O}(N^2 D_h)$, where $N$ indicates the number of causal factors and $D_h$ describes the hidden dimensionality. The transformations in the PeNF are linear. Thus the total complexity is $\mathcal{O}(T \log T) + \mathcal{O}(N^2 D_h)$.


\section{Experiment Settings}\label{app:CaPulse_setting}
We implement CaPulse and baselines with PyTorch 1.10.2 on one NVIDIA A100. We follow the setting of previous works~\citep{dai2022graphaugmented, zhou2023detecting} to split datasets by 60\% for training, 20\% for validation, and 20\% for testing. The sequen lenghth of the input time series are set to 60. Our model is trained using Adam optimizer~\citep{kingma2014adam} with a learning rate of 0.001. We implement the high-frequency threshold $k_h = 25\% T$ in causal intervention in Eq.~\ref{eq:aug} and the amplitude of intervention $\sigma$ we search over \{0.01, 0.1, 1, 2, 10\}. For the hidden dimension $D_h$, we conduct a grid search over \{8, 16, 32, 64\}. For the number of layers and blocks, we test it from 1 to 5. The balance coefficients in the loss function $\alpha$ and $\beta$ are searched over \{0.001, 0.1, 0.05, 0.1, 0.2\}. We outline the optimal hyperparameter configurations used for CaPulse across different datasets:
\begin{itemize}[leftmargin=*]
    \item \textbf{Cloud-B:} We set the hidden size to 32, the number of blocks to 2, and the number of layers to 2. The balancing coefficients for the mutual information loss, $\alpha$, and $\beta$, were both set to 0.1, ensuring an appropriate trade-off between different loss components.
    
    \item \textbf{Cloud-S:} For Cloud-S, the hidden size is set to 32, with 2 blocks and 1 layer. The mutual information loss coefficients $\alpha$ and $\beta$ were set to 0.01 and 0.1, respectively.
    
    \item \textbf{Cloud-Y:} In this case, the hidden size was set to 32, the number of blocks to 3, and the number of layers to 1. The mutual information loss coefficients $\alpha$ and $\beta$ were both set to 0.1.
    
    \item \textbf{WADI:} The WADI dataset used a hidden size of 32, with 1 block and 1 layer. The mutual information loss coefficients $\alpha$ and $\beta$ were both set to 0.05.
    
    \item \textbf{PSM:} For PSM, we configured the model with a hidden size of 32, 1 block, and 1 layer. The mutual information loss coefficients were set to $\alpha = 0.1$ and $\beta = 0.1$.
    
    \item \textbf{SMD:} The model for SMD was also configured with a hidden size of 32, 5 blocks, and 2 layers. The balancing coefficients for the mutual information loss were both set to 0.01.
    
    \item \textbf{MSL:} For the MSL dataset, we set the hidden size to 32, the number of blocks to 1, and the number of layers to 1. The mutual information loss coefficients $\alpha$ and $\beta$ were both set to 0.1.
\end{itemize}

\section{Details of Datasets}\label{app:dataset}

We evaluate the proposed model on seven real-world datasets from different domains, including five commonly used public datasets for TSAD - MSL (Mars Science
Laboratory rover)~\citep{hundman2018detecting}, SMD (Server Machine Dataset)~\citep{su2019robust}, PSM (Pooled Server Metrics)~\citep{abdulaal2021practical}, WADI (Water Distribution)~\citep{ahmed2017wadi} - and three cloud computing platform datasets, namely Cloud-B, Cloud-S, and Cloud-Y, collected by our company~\footnote{Company details temporally omitted for anonymity.}. Each dataset consists of multivariate monitoring metrics recorded at different time points from a single instance (i.e., virtual machine). These metrics include factors such as the number of slow-running tasks, CPU usage, and memory consumption. The labels indicate whether any issues occurred in the monitored instance.

\begin{table}[h]
  \centering
  \footnotesize
  \caption{Detail of datasets. \# Train/Val/Test: the number of training/validation/test samples.}\label{tab:dataset}
\begin{tabular}{lccccc}
\shline
\textbf{Dataset} & \textbf{\# Dims} & \textbf{\# Train} & \textbf{\# Val} & \textbf{\# Test} & \textbf{Anomaly Rate (\%)} \\
\hline
\textbf{Cloud-B}          & 6                & 14,604            & 4,868           & 4,869            & 5.649                      \\
\textbf{Cloud-S}          & 6                & 14,604            & 4,868           & 4,869            & 4.453                      \\
\textbf{Cloud-Y}            & 6                & 14,604            & 4,868           & 4,869            & 2.703                      \\
\textbf{WADI}             & 123              & 103,680           & 34,560          & 34,561           & 5.774                      \\
\textbf{PSM}              & 25               & 52,704            & 17,568          & 17,569           & 27.756                     \\
\textbf{SMD}              & 38               & 14,224            & 4,741           & 4,742            & 3.037                      \\
\textbf{MSL}              & 55               & 44,237            & 14,745          & 14,746           & 10.533                     \\
\shline
\end{tabular}
\end{table}

\section{Details of Baselines}\label{app:baseline}

We opted to include a selection of widely-used cutting-edge methods for comparative evaluation. We describe these baselines used in our experiments {and their settings} as follows. {We use the same setting for all datasets.}
\begin{itemize}[leftmargin=*]
    \item \textbf{DeepSVDD}~\citep{ruff2018deep} Deep Support Vector Data Description (DeepSVDD) is a deep learning-based anomaly detection method that minimizes the volume of a hypersphere enclosing the normal data in the latent space. We utilize the publicly available implementation\footnote{https://github.com/lukasruff/Deep-SVDD-PyTorch} for our experiments. {The hidden dimension is set to 64, the number of layers are set to 2.}
    
    \item \textbf{ALOCC}~\citep{sabokrou2020deep}: Adversarially Learned One-Class Classifier (ALOCC) leverages GANs to learn compact representations of normal data for detecting anomalies. We use the official implementation\footnote{https://github.com/khalooei/ALOCC-CVPR2018} provided by the authors. {We set the hidden dimension to 64 and the number of layers to 2.}
    
    \item \textbf{DROCC}~\citep{goyal2020drocc}: Deep Robust One-Class Classification (DROCC) is a method that generates adversarial perturbations around the normal data to improve robustness for anomaly detection. The authors’ code\footnote{https://github.com/microsoft/EdgeML/tree/master/pytorch} is employed for our experiments. {The model uses a hidden dimension of 64 and consists of 2 layers. We set gamma (parameter to vary projection) to 2 and lamda (weight given to the adversarial loss) to 0.0001.}
    
    \item \textbf{DeepSAD}~\citep{ruff2019deep}: Deep Semi-Supervised Anomaly Detection (DeepSAD) builds on DeepSVDD by incorporating labeled anomalies during training, aiming for improved detection of rare anomalies. We adopt the publicly released code\footnote{https://github.com/lukasruff/Deep-SAD-PyTorch} for our analysis. {A hidden dimension of 64 is employed, with the number of layers fixed at 2.}
    
    \item \textbf{DAGMM}~\citep{zong2018deep}: Deep Autoencoding Gaussian Mixture Model (DAGMM) jointly optimizes a deep autoencoder and a Gaussian mixture model to detect anomalies based on reconstruction errors and low-dimensional latent representations. We leverage the code\footnote{https://github.com/danieltan07/dagmm} shared by the authors. {The hidden size is defined as 64, and the network is composed of 2 layers.}
    
    \item \textbf{USAD}~\citep{audibert2020usad}: UnSupervised Anomaly Detection (USAD) is an unsupervised method designed for multivariate time series, using autoencoders to learn normal patterns and detect anomalies. The authors’ implementation\footnote{https://github.com/manigalati/usad} is employed in our study. {For this configuration, the hidden dimension is 64, and the model has 2 layers. $\alpha$ and $\beta$ are both set to 0.5.}

    \item \textbf{AnomalyTransformer}~\citep{xu2022anomaly}: Anomaly Transformer introduces a novel approach for unsupervised time series anomaly detection by leveraging an Association Discrepancy criterion, an innovative Anomaly-Attention mechanism, and a minimax strategy to enhance the differentiation between normal and abnormal patterns. The official code\footnote{https://github.com/thuml/Anomaly-Transformer} is employed for our experiments. {The window size is set to 60, the number of attention heads is 8, and the feedforward network dimension is 512.}
    
    \item \textbf{GANF}~\citep{dai2022graphaugmented}: Graph-Augmented Normalizing Flows (GANF) leverages normalizing flows conditioned on a graph neural network for unsupervised anomaly detection in multivariate time series. We utilize the official code\footnote{https://github.com/EnyanDai/GANF} for our experiments. {We configure the hidden size to 32 and set the number of blocks to 1.}
    
    \item \textbf{MTGFlow}~\citep{zhou2023detecting}: MTGFlow uses entity-aware normalizing flows to capture multiscale dependencies in time series data for anomaly detection. We rely on the authors’ released code\footnote{https://github.com/zqhang/MTGFLOW} for conducting our experiments. {The setup involves a hidden dimension of 32 and a total of 2 layers.}

\end{itemize}


%
%
\section{More Experimental results}
\subsection{Statistical Significance Analysis}\label{app:significance_analysis}
To evaluate whether the performance improvements of CaPulse over existing baselines are statistically significant, we conduct the Wilcoxon signed-rank test~\citep{conover1999practical} on our main baseline results in Table~\ref{tab:results}. The resulting $p$-values and significance levels are summarized in Table~\ref{tab:wilcoxon}. As shown in the table, CaPulse achieves statistically significant improvements over most baselines, especially compared to DeepSVDD, ALOCC, DeepSAD, DAGMM, and USAD, with $p$-values below 0.05. This provides further evidence of the effectiveness of our method.

\begin{table}[h]
\centering
\footnotesize
\caption{Wilcoxon signed-rank test results comparing \textit{CaPulse} with baselines. Significance level: * $p < 0.05$, ** $p < 0.01$.}
\label{tab:wilcoxon}
\begin{tabular}{lcc}
\shline
\textbf{Baseline} & \textbf{$p$-value} & \textbf{Significance Level} \\
\midrule
\textbf{DeepSVDD} & 0.0219 & ** \\
\textbf{ALOCC} & 0.0383 & ** \\
\textbf{DROCC} & 0.0959 & * \\
\textbf{DeepSAD} & 0.0248 & ** \\
\textbf{DAGMM} & 0.0338 & ** \\
\textbf{USAD} & 0.0294 & ** \\
\textbf{AnomalyTransformer} & 0.0734 & * \\
\textbf{TimesNet} & 0.1223 & \\
\textbf{DualTF} & 0.1363 & \\
\textbf{GANF} & 0.1164 & \\
\textbf{MTGFLOW} & 0.1444 & \\
\shline
\end{tabular}
\end{table}

\subsection{Ablation studies}\label{app:ablation}
To further demonstrate the generalizability of our approach, we conducted ablation studies on two additional datasets beyond those described in Section~\ref{sec:ablation}. The results of these experiments are presented in Table~\ref{tab:app_ablation}. The results show that removing any single component leads to noticeable performance drops, ranging from 3.46\% to 4.1\% on Cloud-B, 3.59\% to 3.98\% on PSM, and 4.58\% to 6.87\% on WADI. In contrast, the full CaPulse model consistently achieves the highest performance.

\begin{table}[h]
\centering
\footnotesize
\caption{Variant results on the Cloud-B, PSM, and WADI datasets.}\label{tab:app_ablation}
\begin{tabular}{llll}
\shline
\textbf{Dataset}  & \textbf{Cloud-B}                       & \textbf{PSM}                           & \textbf{WADI} \\
\hline
\textbf{w/o CI}   & 0.888 $\pm$ 0.002 (\blue{$\downarrow$4.1\%})  & \secondcolor 0.726 $\pm$ 0.009 (\blue{$\downarrow$3.59\%}) & 0.775 $\pm$ 0.027 (\blue{$\downarrow$6.63\%}) \\
\textbf{w/o ICM}  & 0.889 $\pm$ 0.006 (\blue{$\downarrow$4\%})    & 0.725 $\pm$ 0.002 (\blue{$\downarrow$3.72\%})              & 0.792 $\pm$ 0.031 (\blue{$\downarrow$4.58\%}) \\
\textbf{w/o Attn} & 0.891 $\pm$ 0.002 (\blue{$\downarrow$3.78\%}) & 0.723 $\pm$ 0.010 (\blue{$\downarrow$3.98\%})              & 0.773 $\pm$ 0.028 (\blue{$\downarrow$6.87\%}) \\
\textbf{w/o GP}   & \secondcolor 0.894 $\pm$ 0.001 (\blue{$\downarrow$3.46\%}) & 0.725 $\pm$ 0.009 (\blue{$\downarrow$3.72\%}) & 0.774 $\pm$ 0.043 (\blue{$\downarrow$6.75\%}) \\
\hline
\textbf{CaPulse}  & \bestcolor 0.926 $\pm$ 0.007                      & \bestcolor 0.753 $\pm$ 0.042           & \bestcolor 0.830 $\pm$ 0.029 \\
\shline
\end{tabular}
\end{table}

\subsection{Augmentation methods for causal intervention}
In Section~\ref{sec:causal_treatments}, we conduct causal interventions by injecting Gaussian noise into the less significant high-frequency components of the input data, aiming to simulate real-world disturbances. 
To assess the robustness of this design, we conducted additional experiment with other augmentation strategies. The ROC results on two datasets PSM and SMD are reported in Table~\ref{tab:aug_results}.
Specifically, \textbf{HighFreq} denotes our original approach of adding noise to high-frequency components, while \textbf{LowFreq} refers to noise added to low-frequency components. \textbf{Shift} represents a temporal shift of the input time series by 20 time steps. The $+$ symbol indicates the combination of multiple augmentation methods.
To better capture complex noise scenarios, we also experimented with Laplace-distributed noise in addition to Gaussian noise. Laplace noise introduces heavy-tailed and asymmetric variations. The type of noise used is indicated in brackets.

\begin{table}[h]
  \centering
  \footnotesize
  \caption{5-run results for different augmentation methods to implement causal intervention.}\label{tab:aug_results}
\begin{tabular}{lcc}
\shline
\textbf{Augmentation   Method}      & \textbf{PSM}  & \textbf{SMD}  \\
\hline
\textbf{HighFreq} (Gaussian)                 & \bestcolor \textbf{0.753} ± 0.042 & \bestcolor \textbf{0.906} ± 0.009 \\
\textbf{LowFreq}   (Gaussian)                 &  0.743 ± 0.015 & 0.902 ± 0.007 \\
\textbf{HighFreq} (Laplace)                                 & \secondcolor \underline{0.747} ± 0.011          & 0.905 ± 0.006          \\
\textbf{LowFreq} (Laplace)                                  & 0.728 ± 0.015          & 0.893 ± 0.007          \\
\textbf{Shift}                      & 0.728 ± 0.011 & 0.885 ± 0.022 \\
\textbf{HighFreq} (Gaussian) + \textbf{LowFreq}  (Gaussian)       & 0.725 ± 0.009 & \secondcolor \underline{0.905} ± 0.005 \\
\textbf{HighFreq} (Gaussian) + \textbf{Shift}           & 0.727 ± 0.011 & 0.884 ± 0.021 \\
\textbf{LowFreq} (Gaussian) + \textbf{Shift}            & 0.725 ± 0.008 & 0.881 ± 0.018 \\
\textbf{HighFreq} (Gaussian)+ \textbf{LowFreq} (Gaussian) + \textbf{Shift} & 0.729 ± 0.014 & 0.874 ± 0.010\\
\shline
\end{tabular}
\end{table}

\paragraph{Analysis of perturbation location.} 
We first focus on Gaussian-based interventions for perturbation location analysis. Among all methods, \textbf{HighFreq} (Gaussian) consistently yields the best performance on both datasets (PSM: 0.753, SMD: 0.906), indicating that injecting noise into high-frequency components is most effective for simulating realistic disturbances and enhancing anomaly detection. 
\textbf{LowFreq} (Gaussian) also performs reasonably well but slightly lags behind, suggesting that perturbing long-term trends contributes less to useful supervision.
\textbf{Shift}-based interventions show the lowest performance, implying limited utility in mimicking causal disturbances.
Furthermore, combining multiple augmentation methods (e.g., HighFreq + LowFreq or + Shift) does not lead to additional gains and sometimes degrades performance, likely due to over-complicated or conflicting perturbations.

\paragraph{Comparision of noise type.}
We then compare the effect of different noise distributions (Gaussian vs. Laplace) under the same injection strategy. Table~\ref{tab:aug_noise_significance} reports the statistical significance when comparing Gaussian and Laplace noise.
According to the table, the differences in AUROC between Gaussian and Laplace noise are not statistically significant in most settings ($p > 0.05$), with only one marginal case (\textbf{SMD} \textbf{LowFreq}, $p = 0.0403 < 0.05$). This suggests that CaPulse remains robust under diverse noise distributions and is able to consistently isolate meaningful causal factors even in more challenging, non-Gaussian conditions.

\begin{table}[h]
\footnotesize
  \centering
  \caption{Statistical significance analysis comparing Gaussian and Laplace noise injection. }\label{tab:aug_noise_significance}
\begin{tabular}{lccc}
\shline
                                   & \textbf{P-value} & $<0.05$ & $<0.01$ \\\hline
\textbf{PSM} \textbf{HighFreq} (Gaussian vs Laplace) & 0.4486  & \ding{55}     & \ding{55}    \\
\textbf{PSM} \textbf{LowFreq} (Gaussian vs Laplace)  & 0.1281  & \ding{55}     & \ding{55}     \\
\textbf{SMD} \textbf{HighFreq} (Gaussian vs Laplace) & 0.2703  & \ding{55}     & \ding{55}     \\
\textbf{SMD} \textbf{LowFreq} (Gaussian vs Laplace)  & 0.0403  & \ding{51}    &  \ding{55} \\\shline   
\end{tabular}
\end{table}

\begin{figure}[H]
    \centering
    \includegraphics[width=0.8\textwidth]{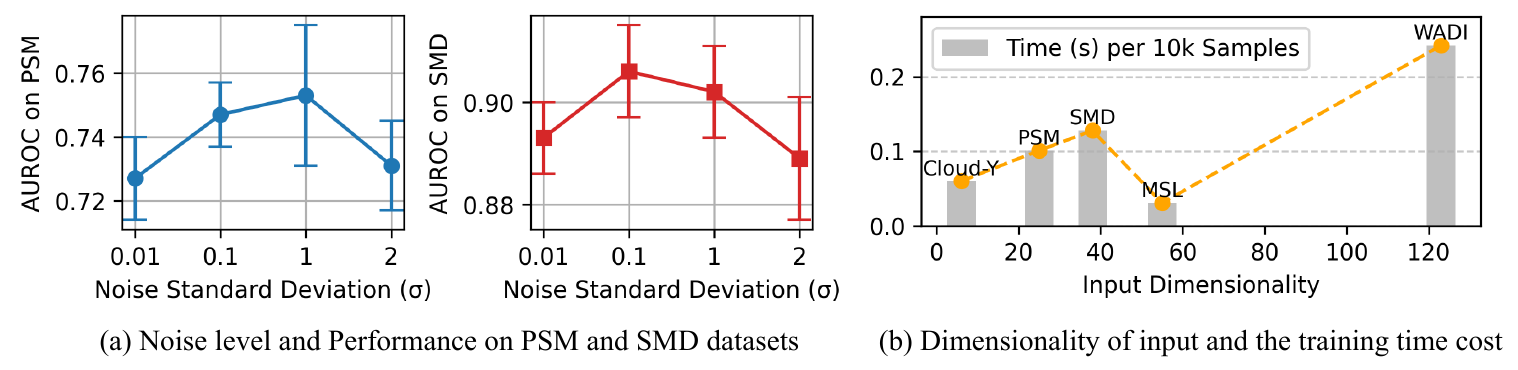}\label{fig:noise_time}
    \caption{(a) AUROC under varying noise levels on PSM and SMD. Error bars show standard deviation. (b) Training time (per $10k$ samples) vs. input dimensionality across datasets. }\end{figure}

\paragraph{Sensitivity of noise level.}
To evaluate the robustness of our method to different noise magnitudes during causal intervention, we conduct a sensitivity analysis by varying the standard deviation \(\sigma\) of the Gaussian noise injected into the high-frequency components. Figure~\ref{fig:noise_time}a presents the AUROC performance on the PSM and SMD datasets under \(\sigma \in \{0.01, 0.1, 1, 2\}\). We observe that the performance improves as the noise level increases from \(\sigma = 0.01\) to \(\sigma = 0.1\) or \(1.0\), reaching the peak performance at moderate noise levels. Specifically, \(\sigma = 0.1\) yields the best AUROC on SMD, while \(\sigma = 1.0\) slightly outperforms others on PSM. When the noise level becomes too large (\(\sigma = 2.0\)), the performance drops, likely due to excessive perturbation that overwhelms meaningful temporal patterns. These results suggest that our method is robust to a reasonable range of noise levels, and moderate noise magnitudes are most effective for simulating realistic disturbances without distorting the underlying causal structure.

\begin{wrapfigure}{R}{0.35\textwidth}
\vspace{-0.8em}
  \includegraphics[width=\linewidth]{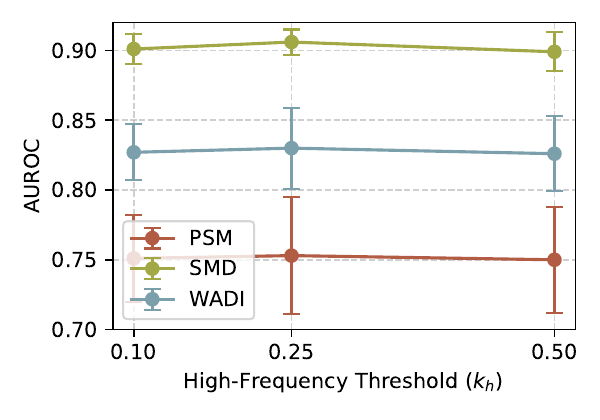}
  \vspace{-2em}
\caption{Sensitivity to high-frequency threshold $k_h$ across datasets.}\label{fig:high_freq}
  \vspace{-0.8em}
\end{wrapfigure}

\paragraph{Clarification on high-frequency threshold.}
The threshold for high-frequency components in our method is set to $k_h = 0.25$, selected via a coarse hyperparameter sweep on validation sets across multiple datasets. This value is globally fixed and remains unchanged across datasets—no dataset-specific tuning is performed.
To evaluate the robustness of this setting, we conduct a sensitivity analysis on three representative datasets (PSM, SMD, WADI), as shown in Figure~\ref{fig:high_freq}. The results reveal only marginal AUROC fluctuations across a range of $k_h$ values, suggesting that CaPulse is largely insensitive to this hyperparameter.
This design choice ensures that no a priori knowledge or test-set-specific information is exploited during threshold selection, and that the model maintains generalizability across datasets with diverse spectral characteristics.

\subsection{Efficiency comparison}~\label{app:sec-efficiency}
We compare our method with some classical baselines to demonstrate the methods' efficiency. For theoretical computational complexity, we have discussed in Appendix~\ref{app:complexity}.

\textbf{Training time and input dimension.} We plot training time (per $10k$ samples) against input dimensionality across datasets in Figure~\ref{fig:noise_time}b. The dataset statistics are provided in Table~\ref{tab:dataset}. According to the figure, despite high input dimensions (e.g., WADI with 123 features), the training time remains low ($\leq 0.025s$ per $10k$ samples) and scales nearly linearly with data size (e.g., from SMD to WADI: $14k \rightarrow 103k$ samples, time increases by only $1.88\times$). These results confirm that our method remains efficient even in high-dimensional, large-scale industrial scenarios.

\textbf{Time cost and parameter comparison.} We first compare the time cost and parameter of CaPulse and one of the classical TSAD method Matrix Profile (MP)~\citep{yeh2016matrix}.
Theoretically, the complexity of MP is $\mathcal{O}(T_l^2 \log T_l)$, where $T_l$ represents the total length of the time series (typically, $T_l \gg T$). Thus, MP’s theoretical complexity is higher than that of our approach. We conducted experiments on four datasets and measured the time costs. Note, that we believe that a direct efficiency comparison may be unfair for several reasons:
    1) Methods like MP can only be run on the CPU, while DL methods such as CaPulse can leverage GPU acceleration.
    2) MP operates directly on the test data, which is smaller (about one-third of the training set size), whereas CaPulse is trained on the full training set.
    3) Training epochs vary across datasets and can be adjusted, making the total training time flexible.
Thus, to provide additional context, we also included a comparison with a recent DL-based method, DualTF~\cite{nam2024breaking}. The results are summarized in Table~\ref{tab:app_time_comparison}, where we observe that CaPulse achieves significantly lower time costs per epoch and consistently outperforms both MP and DualTF in ROC scores, demonstrating both efficiency and effectiveness.

\begin{table}[ht]
\footnotesize
\centering
\caption{Comparison of efficiency of methods across datasets. The magnitude of \#Param (the number of parameters) is Kilo. Time is reported in seconds for MP and seconds per epoch for DualTF and CaPulse.}
\label{tab:app_time_comparison}
\begin{tabular}{ccccc}
\shline
\textbf{Dataset} & \textbf{Metric}         & \textbf{MP} & \textbf{DualTF}        & \textbf{CaPulse}                \\ \hline
\textbf{PSM}     & \textbf{\#Param (K)}       & -                      & 4801.6                 & 204.7                           \\
                 & \textbf{Time}          & 25.944                & 2.265 ± 0.356          & 0.533 ± 0.192                   \\
                 & \textbf{ROC-AUC}                 & 0.634                  & 0.727 ± 0.071          & 0.753 ± 0.042                   \\ \hline
\textbf{SMD}     & \textbf{\#Param (K)}       & -                      & 4820                   & 264.7                           \\
                 & \textbf{Time}          & 24.673                 & 0.709 ± 0.385          & 0.182 ± 0.195                   \\
                 & \textbf{ROC-AUC}                 & 0.866                  & 0.796 ± 0.101          & 0.906 ± 0.009                   \\ \hline
\textbf{WADI}    & \textbf{\#Param (K)}       & -                      & 4949.1                 & 342.2                           \\
                 & \textbf{Time}          & 40.428                 & 4.52 ± 0.372           & 2.505 ± 0.197                   \\
                 & \textbf{ROC-AUC}                 & 0.677                  & 0.796 ± 0.030          & 0.830 ± 0.029                   \\ \hline
\textbf{SWaT}    & \textbf{\#Param (K)}       & -                      & 4840.5                 & 242.4                           \\
                 & \textbf{Time}          & 43.065                 & 11.244 ± 0.34          & 3.613 ± 0.243                   \\
                 & \textbf{ROC-AUC }                & 0.600                  & 0.769 ± 0.019          & 0.782 ± 0.004                   \\ \shline
\end{tabular}
\end{table}

\subsection{Additional classical baselines}
To further compare our method with classical baselines, we have compared it with three additional baselines, i.e., MP, KNN, and K-means. The results are shown in Table~\ref{tab:baselines}, which demonstrate that CaPulse consistently achieves superior ROC scores compared to classical methods, reinforcing its robustness and accuracy in detecting anomalies across diverse datasets.

\begin{table}[ht]
\footnotesize
\centering
\caption{Comparison with classical baselines.}
\label{tab:baselines}
\begin{tabular}{ccccc}
\shline
\textbf{}               & \textbf{SWaT}        & \textbf{WADI}        & \textbf{PSM}         & \textbf{SMD}         \\ \hline
\textbf{MP}  & 0.600                & 0.677                & 0.634                & 0.866                \\
\textbf{KNN}            & 0.716                & 0.815                & 0.654                & 0.496                \\ 
\textbf{K-means}        & 0.560                & 0.639                & 0.535                & 0.692                \\ 
\textbf{CaPulse}        & 0.782 ± 0.004        & 0.830 ± 0.029        & 0.753 ± 0.042        & 0.906 ± 0.009        \\ \shline
\end{tabular}
\end{table}

\section{More Discussions}\label{app:discussion}

\subsection{The role of causality in this work}
Causality in our work serves as a \textbf{design principle}, not as a target of inference. That is, we do \textbf{not} perform causal discovery or identify causal relationships among observed variables. Instead, we assume the existence of \textbf{latent} causal factors and use causal theory to inform model design.

Our approach is grounded in established causal theory (Section~\ref{sec:causal_view}), and these assumptions directly guide how we structure and train the model (Section~\ref{sec:causal_treatments}). Specifically:
\begin{itemize}[leftmargin=*]
    \item \textbf{Structural Causal Model (SCM)} (Section~\ref{sec:scm}): 
    We model the anomaly generation process using an SCM, distinguishing between latent causal factors ($C$) and non-causal noise ($U$). Instead of directly modeling $P(y \mid X)$, we formulate the problem as learning $P(y \mid do(U), C)$ to capture the true causal drivers of anomalies.
    \item \textbf{Causal Principles} (Section~\ref{sec:causal_backing}): Common Cause Principle assumes that the observed variables share a common latent cause and Independent Causal Mechanisms assumes that the generating mechanisms of different causal factors are mutually independent.
    \item \textbf{Causality-Guided Model Design} (Section~\ref{sec:causal_treatments}): Guided by the above, our model introduces: (1) \textbf{causal intervention} realized through noise injection to enforce independence between $C$ and $U$, 
    and (2) \textbf{a joint independence loss}, implemented via orthogonality constraints to encourage mutual independence among the dimensions of $C$.
    
\end{itemize}

This principled design ensures that the learned representations focus on invariant, causally relevant signals rather than spurious correlations.

\subsection{Applicability of the proposed SCM in real-world scenarios}\label{app:scm}

In Section~\ref{sec:scm}, we introduced a causal perspective on the TSAD task by proposing a Structural Causal Model (SCM), as illustrated in Figure~\ref{fig:scm}b. In the proposed SCM, 
the non-causal factors $\boldsymbol{U}$ and the causal factors $\boldsymbol{C}$ are defined as unobserved latent variables that represent a range of potential influences. Based on whether a factor directly causes $\boldsymbol{y}$ or only affects $\boldsymbol{X}$ without impacting $\boldsymbol{y}$, we can categorize it as either a causal factor $\boldsymbol{C}$ or a non-causal factor $\boldsymbol{U}$. This distinction is therefore flexible and may vary depending on the specific domain or scenario. We acknowledge that real-world environments can be more complex and dynamic than our model assumptions. Nevertheless, we believe that \textit{fundamental} patterns in real-world settings can be effectively abstracted and represented within this SCM framework for TSAD.

To further support this point, in addition to the cloud computing platform example provided in Section~\ref{sec:scm}, we offer another real-world scenario in healthcare. In this context, $\boldsymbol{X}$ could represent biometric data (e.g., heart rate, sleep patterns) collected from wearable devices, with anomalies $\boldsymbol{y}$ indicating potential health issues. Here, $\boldsymbol{U}$ might correspond to environmental factors or background noise that influence the readings in $\boldsymbol{X}$ without signifying genuine bodily anomalies, while $\boldsymbol{C}$ could represent factors such as medication side effects that directly impact both $\boldsymbol{X}$ and $\boldsymbol{y}$. Thus, this adaptability enables our model to accommodate different domains by appropriately classifying factors as causal or non-causal based on their direct or indirect influence on the anomaly outcome.

\subsection{High-frequncey strategy of causal intervention}
In the domains targeted by our benchmark datasets (e.g., industrial operations, cloud systems, and sensor-based monitoring), high-frequency variations are commonly linked to noise, sensor jitter, or random fluctuations, rather than semantically meaningful causes. Therefore, high-frequency perturbation is a reasonable and practical design choice for simulating exogenous interventions under these scenarios. Yet, in some other domains (e.g., financial markets, biomedical signals), high-frequency signals can contain meaningful causal information, and intervention strategies should be adapted accordingly. 

\subsection{Periodocity in dataset}
\begin{figure}[t]
    \centering
    \includegraphics[width=0.8\textwidth]{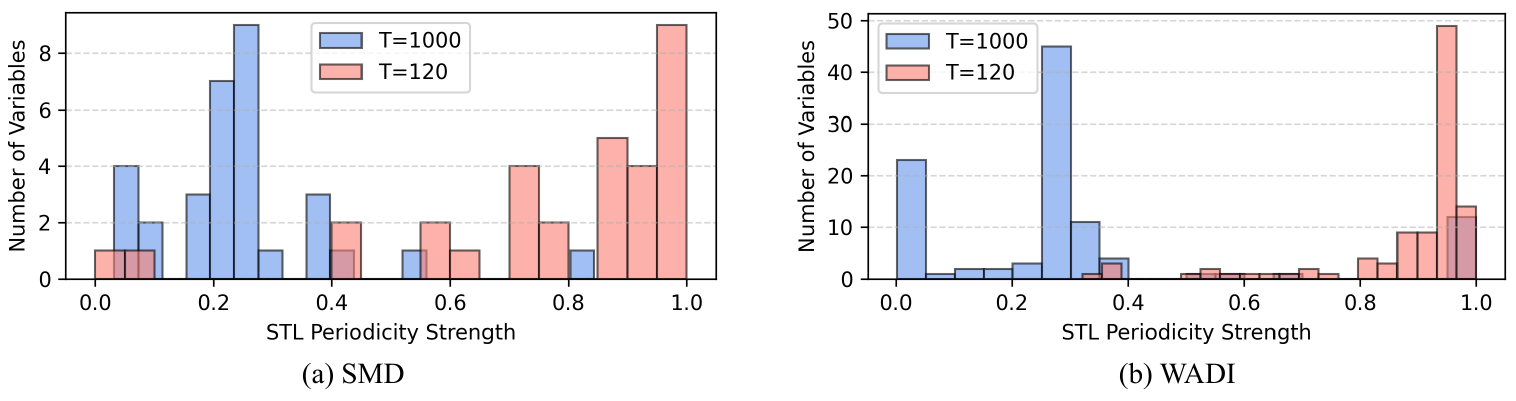}
    \caption{STL periodicity strength distributions for SMD and WADI datasets under different time series lengths (T = 1000 vs. T = 120).}\label{fig:period_dataset}
    \end{figure}

Real-world time series are often non-stationary, and their periodic or seasonal patterns may be local, subtle, or intermittent. Thus some datasets used in our experiments (e.g., WADI, SMD, and MSL) may not exhibit strong and clear long-term periodicity. To assess periodicity strength, we perform STL decomposition and compute the following metric:
\[
F_S = \max\left(0, 1 - \frac{\text{Var}(R_t)}{\text{Var}(S_t + R_t)}\right),
\]
where \( R_t \) and \( S_t \) are the residual and seasonal components, respectively. Figure~\ref{fig:period_dataset} shows the distribution of periodicity strength scores for the SMD and WADI datasets. While most variables demonstrate weak periodicity when using a long window (e.g., 1000 time steps), we observe significantly stronger periodic patterns when using shorter windows (e.g., 120 time steps), suggesting that short-term periodic structures can still be effectively captured.
Note that these datasets are only a subset of our evaluation suite. We intentionally include datasets with varying temporal characteristics to assess the generalizability of our model. Our approach is not designed to rely solely on strong periodic signals but rather to adaptively learn useful temporal structures when present.
Our ablation studies (see Table~\ref{tab:ablation} and Table~\ref{tab:app_ablation}) further demonstrate the utility of modeling global periodic context. When the PC-Mask module is removed (i.e., \textbf{w/o GP}), we observe consistent performance degradation—even on datasets with weak or localized periodicity. This empirically supports the benefit of incorporating global information, regardless of the strength of the underlying periodic signal.
Recent works have specifically explored periodic structure discovery in time series using attention-based or unsupervised mechanisms~\cite{yu2024amortizedperiod,demirel2024unsupervised}. Although these methods target periodicity identification rather than anomaly detection, their approaches may be complementary. In future work, such techniques could potentially enhance components in our framework—e.g., serving as replacements or augmentations to the frequency selection and masking in PaCM—especially under weak or local periodic signals.

Admittedly, recent works have specifically explored periodic structure discovery in time series using attention-based or unsupervised mechanisms~\cite{yu2024amortizedperiod,demirel2024unsupervised}. In future work, such techniques could potentially enhance components in our framework—e.g., serving as replacements or augmentations to the frequency selection and masking in PaCM—especially under weak or local periodic signals.

\subsection{Comparison with related works}
CaPulse addresses key gaps in existing methods for TSAD by introducing a causal and periodicity-aware approach. Density-based TSAD methods such as GANF~\citep{dai2022graphaugmented} and MTGFlow~\citep{xu2023density} lack a causal perspective and fail to account for multi-periodicity in time series data. While forecasting-~\citep{hundman2018detecting,shen2020timeseries} and reconstruction-based models~\citep{su2019robust,audibert2020usad,xu2022anomaly}  improve anomaly detection, they rely solely on statistical patterns without capturing underlying causal processes, making them less robust to noise and dynamic changes. 
In contrast, CaPulse enhances interpretability and robustness by integrating a causal view and density estimation, specifically tailored to handle complex temporal rhythms. 
While causal inference-based methods such as COST~\citep{cost} and CaseQ~\citep{yang2022towards} have shown promise in forecasting and sequential event prediction, they are not designed for anomaly detection. 

Some recent efforts~\citep{qiu2012granger,xing2023gcformer,fu2024generating, han2025root} focuses on causal discovery between observed variables (e.g., inter-variable Granger causality), which differs fundamentally from our goal. 
CaPulse does not attempt to infer variable-to-variable causal graphs; rather, it models latent, unobserved causal factors within an SCM, capturing how hidden mechanisms give rise to anomalies. This focus on latent causal generative processes uniquely positions CaPulse as a TSAD method that is both theoretically grounded and practically robust.
For these related causal-related TSAD works, we now provide a clearer comparison between in Table~\ref{tab:comparison_related_works}. The main distinction lies in the causal modeling philosophy: whereas previous causal-related works aim to learn causal relationships directly from observational data and subsequently perform anomaly detection, our approach injects causality through a predefined SCM and the principle of independent mechanisms, which guide the design of the anomaly detection model. This design avoids potential overfitting and instability associated with learning causal graphs from noisy or limited data. Additionally, our framework explicitly addresses data-centric challenges including label scarcity, data imbalance, and multiple periodicities, which are common in real-world anomaly detection tasks.

\begin{table}[h]
\centering
\footnotesize
\caption{Comparison between prior causal-related TSAD methods and our approach.}\label{tab:comparison_related_works}
\renewcommand{\arraystretch}{1.2}
\begin{tabular}{p{1.5cm}p{3.2cm}p{4cm}p{4cm}}
\shline
\textbf{Method} & \textbf{Causal Modeling Approach} & \textbf{Anomaly Modeling} & \textbf{Applicability} \\
\hline
\textbf{~\cite{qiu2012granger}} & Learn Granger causality graph via L1 regularization & Compute a ``correlation anomaly'' score for each variable to detect deviations from expected causal dependencies & Assumes linear dependencies; suited for industrial systems \\\hline
\textbf{~\cite{yang2022causal}} & Learn modular causal structure from observational data & Estimate conditional distributions based on causal structure to detect violations of normal mechanisms & Supports root cause analysis \\\hline
\textbf{~\cite{xing2023gcformer}} & Explicitly construct Granger graph and use attention masks to model variable dependencies & Detect abnormal shifts in Granger attention weights & Offers interpretability and stronger modeling power \\\hline
\textbf{~\cite{fu2024generating}} & Use deep generative model to discover fine-grained causal graph & Jointly performs prediction and anomaly detection during generation & Suited for climate and extreme weather scenarios \\\hline
\textbf{Our method} & Knowledge-guided predefined SCM based on the principle of independent mechanisms & Detect anomalies as deviations in low-density regions of the causal distribution & General-purpose applicability; injects causality while also addressing label scarcity, data imbalance, and multi-periodicity challenges
\\\shline
\hline
\end{tabular}
\vspace{-1em}
\end{table}

\subsection{Broader impacts}\label{app:sec-impact}
Our work aims to enhance the interpretability and generalizability of TSAD methods by introducing a causality-based framework. This has potential positive impacts in domains where detecting and understanding anomalies is critical, such as cloud systems, urban operations, and infrastructure monitoring. The ability to distinguish between true causes of anomalies and spurious factors may lead to more reliable and actionable decision-making.

\subsection{Limitations \& future directions}\label{app:sec-limitation}
A potential limitation of CaPulse is its reliance on the assumption that anomalies lie in low-density regions, which may not always hold in complex real-world scenarios — for instance, in high-frequency trading data where significant anomalies may cluster in high-density regions during market events or crashes. Future work could explore relaxing these distributional assumptions and incorporating reversible transformations to generate synthetic anomalies.

\end{document}